\documentclass{egpubl}
\pdfoutput=1

\JournalSubmission
\electronicVersion

\PrintedOrElectronic
\usepackage{t1enc,dfadobe}
\usepackage{xfrac}
\usepackage{epstopdf}
\usepackage{amsmath}
\usepackage{multirow}
\usepackage{placeins}
\usepackage{color}
\usepackage[bottom]{footmisc}
\usepackage{needspace}
\usepackage{cite}
	
\title{Reducing Lateral Visual Biases in Displays}

\author[I. Huberman \& R. Fattal]
{Inbar Huberman and Raanan Fattal\\
 Hebrew University of Jerusalem, Israel\\
}

\begin{document}
\newcommand {\img}{I}
\newcommand {\scrmap}{\psi}
\newcommand {\lum}{Y}
\newcommand {\ems}{E}
\newcommand {\blur}{k}
\newcommand {\perc}{R}
\newcommand {\power}{n}
\newcommand {\new}{newImg}
\newcommand {\range}{r}
\newcommand {\f}{f}

\newcommand {\x}{\mathbf{x}}
\newcommand {\y}{\mathbf{y}}

\newcommand {\dotp}[1]{\langle #1 \rangle}
\newcommand {\rl}{\mathbb{R}}

\newcommand {\tr}{{\scriptstyle \top \displaystyle}}
\newcommand {\half}{\scriptstyle  \frac{1}{2} \displaystyle}

\newcommand {\bt}[1]{\textbf{#1}\normalfont}

\newcommand{\naturalimage}[1]{}
\newcommand {\raanan}[1]{\bf{Raanan: #1}\normalfont}
\newcommand {\inbar}[1]{\textbf {#1}}


\maketitle
\begin{abstract}

The human visual system is composed of multiple physiological components that apply multiple mechanisms in order to cope with the rich visual content it encounters. The complexity of this system leads to non-trivial relations between what we see and what we perceive, and in particular, between the raw intensities of an image that we display and the ones we perceive where various visual biases and illusions are introduced. In this paper we describe a method for reducing a large class of biases related to the lateral inhibition mechanism in the human retina where neurons suppress the activity of neighboring receptors. Among these biases are the well-known Mach bands and halos that appear around smooth and sharp image gradients as well as the appearance of false contrasts between identical regions. The new method removes these visual biases by computing an image that contains counter biases such that when this \emph{laterally-compensated} image is viewed on a display, the inserted biases cancel the ones created in the retina.


User study results confirm the usefulness of the new approach for displaying various classes of images, visualizing physical data more faithfully, and improving the ability to perceive constancy in brightness.

\begin{classification} 
\CCScat{Computer Graphics}{I.2.10}{Vision and Scene Understanding}{Perceptual reasoning}
\end{classification}

\end{abstract}

\section{Introduction}
\label{sec:introduction}

The human visual system is composed of multiple physiological components that apply different mechanisms in order to cope with the rich visual data that we encounter in our daily life. Consequently, there is a highly non-trivial relation between the way an image is perceived and its raw pixel intensities. This complexity leads to different types of visual biases and optical illusions.

\begin{figure}[t]
\centering
\includegraphics[width=3in]{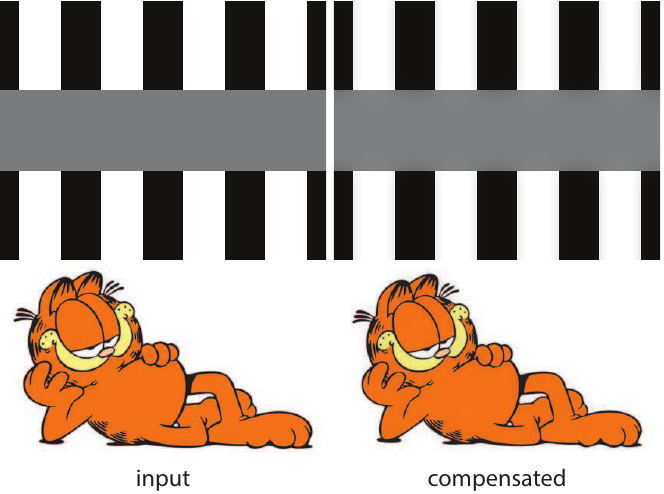}
\caption{\label{fig:teaser}Laterally-compensated images produced by the proposed method. The gray stripe in the Stripes Pattern appears more uniform in the compensated image. Similarly, the perceived halos around the Garfield's outline are reduced in our output image.}
\vspace{-0.3in}
\end{figure}

One of the most common biases is the sharpening of the spatial profile of the response to localized stimuli. This effect is attributed to the \emph{lateral inhibition} mechanism in which stimulated neurons suppress the activity of nearby cells~\cite{von67}. While the differential nature of this process increases the perceived contrast and sharpness, it also leads to various visual artifacts documented in the psychophysical literature. These biases include the well-known Mach bands~\cite{Ratliff65} in which false bright and dark peaks appear around intensity gradients, the introduction of halos around strong image edges~\cite{ghosh06}, as well as the formation of false contrasts, a phenomenon known as simultaneous contrast~\cite{adelson99}. Figures~\ref{fig:teaser} and~\ref{fig:examples} show example images in which some of these effects occur.

While we are accustomed to perceive these inconsistencies on a regular basis, there are various situations in which their elimination is beneficial. In general, the process of imaging consists of converting some physical measurement or computed quantity into pixel intensity with the intent of conveying the actual values. For example, in radiology, tissues are visualized by mapping the amount of absorbed X-ray into gray-level values. This conversion is typically computed pointwise, without taking into account lateral perceptual considerations. While the effects due to lateral inhibition are typically subtle, this straightforward visualization pipeline may lead to misinterpretations of the displayed data and compromise the diagnosis, see~\cite{daff77,nielsen01}. A similar problem is known to occur in observational astronomy where false halos appear in stellar images~\cite{Icke74}. Technical drawings, maps, architectural plans, cartoons, line art, game-graphics, and logos are additional media affected by the lateral biases and may benefit from their elimination. \naturalimage{Even though the human visual system is adjusted to natural scenery and lateral inhibition effects are known to improve the acuity, we show cases in which the removal of excess sharpening yields visually-pleasing photographs that do not contain harsh edges and halos.}

In this paper we describe a new method for generating \emph{laterally-compensated} images containing counter biases, specifically designed to cancel the lateral inhibition effects created when viewed on a display. The method uses an image perception model that accounts for the compresive conversion of the light absorbed by the cone cells to neural excitation and the inhibitory coupling between nearby neurons. We use the well-known Hartline and Ratliff~\shortcite{hartline57} neural network to model the latter. By including and excluding this component we predict the lateral effects that we wish to cancel and compute a target perceived image that does not contain these effects. Thus, we solve an inverse problem where we search for an image that, when viewed on a display and lateral inhibition takes place, appears as some given target image that we wish to display \emph{without} these effects.

The Hartline and Ratliff model is defined over to an achromatic stimulus. Applying it to the different chromatic channels independently produces spurious chromatic halos when processing color images. We extend the method to color images by deriving a lateral inhibition model, based on the Hartline and Ratliff formula, that accounts for the different types of cone cells in the human retina. By assuming chromatically-blind inhibition interactions we avoid the chromatic halos and obtain an effective lateral compensation.

The resulting procedure consists of very simple operations that make it computationally light and allow its integration into the display drivers. Thus, similarly to display color calibration, our technique provides a displaying mode which is free from lateral inhibition effects.

In order to obtain an effective procedure we conducted a user-study in which we estimated the optimal values of various model parameters. We also set up experiments that measure the method's dependency on the viewing distance and moderate changes in the display brightness. In Section~\ref{sec:results} we report the results of these tests as well as ones that demonstrate the effectiveness of our approach in reducing various biases. We also report tests that confirm the method's ability to improve the comprehension of imaged values and perception of brightness constancy.


\section{Related Work}
\label{sec:prev_work}

\begin{figure}[t]
\centering
\includegraphics[width=3in]{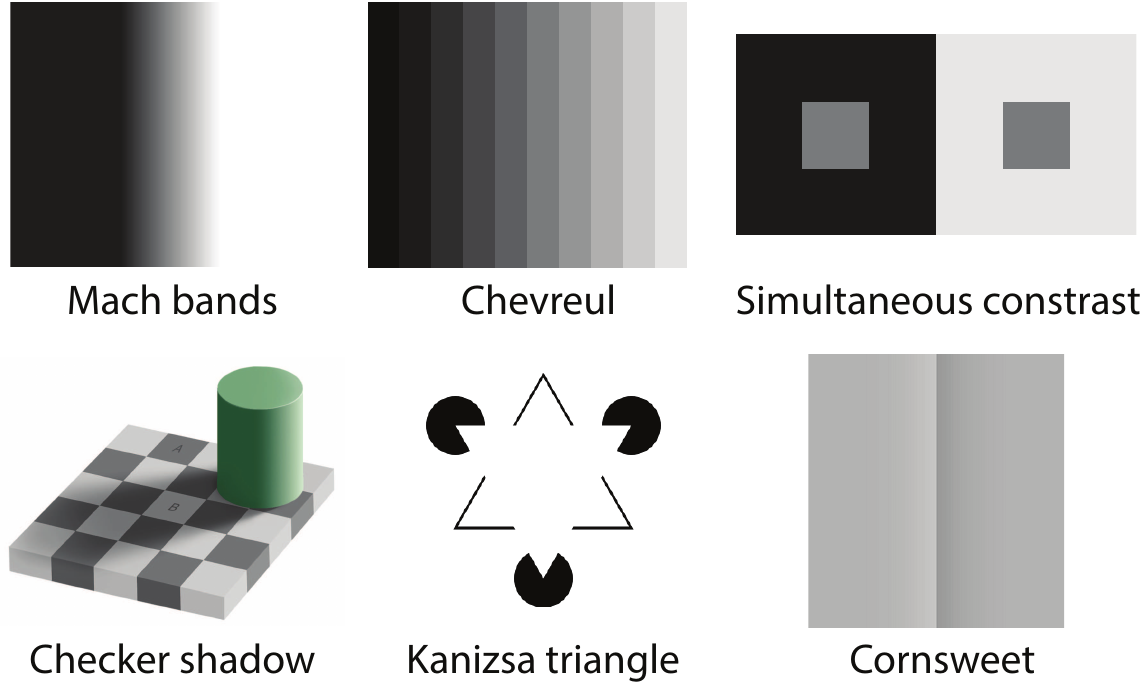}
\caption{\label{fig:examples}Example optical illusions. False brightness variations appear within each fixed gray-scale column in the Chevreul image. The two identical gray squares appear different in the Simultaneous contrast image. Brightness over- and under-shoots appear around the soft gradient in the Mach bands example. In case of the Checkers image the squares titled A and B are identical although they do not appear so. Similarly, the two identical sides of the Cornsweet image appear different. In the Kanizsa triangle a non-extant white triangle emerges. The top three illusions are associated with the lateral inhibition mechanism.}
\vspace{-0.15in}
\end{figure}

\bt{Research in Perception.} Studies of the human visual system suggest that it processes information at different levels of sophistication~\cite{adelson99}. The lowest level takes place in the eye which adapts to the level of incoming light and where retinal lateral inhibition takes place by the center-surround receptive fields of bipolar and ganglion cells. Hartline and Ratliff~\shortcite{hartline57} model the lateral inhibition using a neural network containing inhibitory coupling between nearby neurons. Barlow and Lange~\shortcite{Barlow74} suggest a non-linear variant of this model in which the inhibition depends on the level of excitation. Lateral inhibition is associated with various optical illusions including the Mach bands~\cite{Ratliff65,ghosh06} and the simultaneous contrast~\cite{adelson99} shown in Figure~\ref{fig:examples}. Kingdom and Moulden~\shortcite{kingdom88} propose a contrast sensitivity function specialized for predicting the apparent contrast in Cornsweet-type illusions. Unlike lightness-integration models this edge-detector is valid for arbitrary contrast level. Lateral inhibition effects are typically studied and demonstrated in achromatic stimuli. Tsofe et al.~\shortcite{Spitzer09} discuss related effects occurring in chromatic bands.

The representation and processing of surfaces, contours and object grouping are often attributed to higher-level mechanisms in the visual system. For example, Pessoa et al.~\shortcite{Pessoa88} explain the Cornsweet illusion using the filling-in theory where various visual cues propagate across the image until edges are reached. The Retinex theory, by Land and McCann~\shortcite{land71}, explains the ability to determine the color of a surface (reflectance) and its illumination based on the magnitude of the spatial variation of the two functions. Cognitive processes that incorporate prior knowledge over objects shape and material are attributed to a higher level of the visual system~\cite{gilchrist80,hochberg54}. Figure~\ref{fig:examples} shows example illusions attributed to the different levels of the visual system. In this work, we only account for the low-level lateral inhibition mechanism and hence the method we develop is limited to dealing with visual effects that are associated with this mechanism.

\bt{Applications in Computer Graphics.} Many tone-mapping operators were developed by taking into account various aspects of the human visual system. 
Ledda et al.~\shortcite{ledda04} use the physiological adaptation model of Naka and Rushton~\shortcite{naka66} for
estimating the perceived lightness to derive a tone-mapping. 
Akyuz and Reinhard~\shortcite{Akyuz08} evaluate tone-mapping techniques based on contrast preservation of the Cornsweet illusion image.

A different line of works uses contrast enhancement to modify various perceptual aspects of an image. Luft et al.~\shortcite{Luft06} improve the perception of complex scenes by introducing additional depth cues through the enhancement of contrast, color, and other image parameters. Ritschel et al.~\shortcite{Ritschel08} extend this work to 3D scenes.
Caselles et al.~\shortcite{caselles11} introduce contrast enhancement into tone-mapping operator as an extra stage of visual adaptation. Finally, Trentacoste et al.~\shortcite{Trentacoste12} investigate the perception of contrast enhancement operations and characterize the circumstances in which these operations result in apparent halo artifact. In respect to this close-related work we note that our work focuses on naturally-occurring halo effects in the retina and their cancellation.

In a work, inspiring our own, Kim et al.~\shortcite{kim11} model the influence that edge smoothness has over the perceived lightness and colorfulness. Given an image with a known amount of blur, Kim et al. use their model to restore the color appearance prior to blurring.

Mittelstädt et al.~\shortcite{MSK14} describe a closely related method that compensates for psychological color effects based on the CIECAM02 appearance model. This work focuses on defining the optimization goal and the corresponding perceptual metrics and is demonstrated for data visualization. In our work we derive the perceived image model based on a specific visual mechanism, namely the lateral-inhibition. We demonstrate the reduction of various visual illusions, which were related to this mechanism in the psycho-physical literature. Moreover, we extend the model to color images which enables us to avoid chromatic halos created when processing each color band independently. Beyond its ability to predict the visual biases we wish to eliminate, unlike the non-linear optimization involved in~\shortcite{MSK14}, the linear perceived image model we derive is designed such that it can be efficiently inverted. Mittelstädt and Keim~\shortcite{MSK15} reduce the complexity and running time of~\shortcite{MSK14} by using an efficient optimization and surrogate models.

\vspace{-0.1in}

\section{New Method}
\label{sec:method}
\vspace{-0.05in}

The perceived-image model that we use consists of two well-known components describing the conversion of light to neural excitation and the lateral neural interactions that take place in the retina. We describe these components in Section~\ref{sec:model} in order to establish the quantities that we use later, in Section~\ref{sec:contribution}, to derive our procedure for computing laterally-compensated images. In Section~\ref{sec:pipeline} we extend this perceived-image model to account for the different types of cone cells in the human retina. 
Finally, in Section~\ref{sec:details} we describe a simple modification to our computational pipeline which preserves the fine image details in the compensation process.
\vspace{-0.05in}
\vspace{-0.05in}

\subsection{Perceived-Image Model}
\label{sec:model}

When absorbing light of intensity $y$, the cone photoreceptors produce a graded excitation $e$ which is typically modeled by a compressive function $\phi$,
\begin{equation}
\label{eq:curve}
e = \phi(y).
\end{equation}
The psychophysiological literature offers different functions that model this relation, namely using a logarithmic function (Fechner-Weber)~\cite{fuortes59,fechner66,palmer99}, a power-law relation~\cite{Stevens67} and an S-curve~\cite{naka66}. See~\cite{xie89} for a more elaborate discussion regarding these possibilities and their relation. In our implementation we use the Fechner-Weber compression, $\phi(y)=\log(y)$. Since we focus on the foveal vision in bright visual conditions, we neglect the rod cells excitation.

The photoreceptor excitation $e$ is then transmitted to the bipolar cells either by direct connections or via the horizontal cells which are connected to a set of photoreceptors that surround the central group of directly-connected receptors. These two pathways form two opposite components in the receptive field of the bipolar cells: (i) a central receptive field where the cell reacts excitatorially to the stimulus arriving directly from the photoreceptors, and (ii) a surrounding field to which the cell reacts inhibitorially to stimulus arriving via the horizontal cells. See Palmer~\shortcite{palmer99} for more details.

\begin{figure}[t]
\centering
\includegraphics[width=3in]{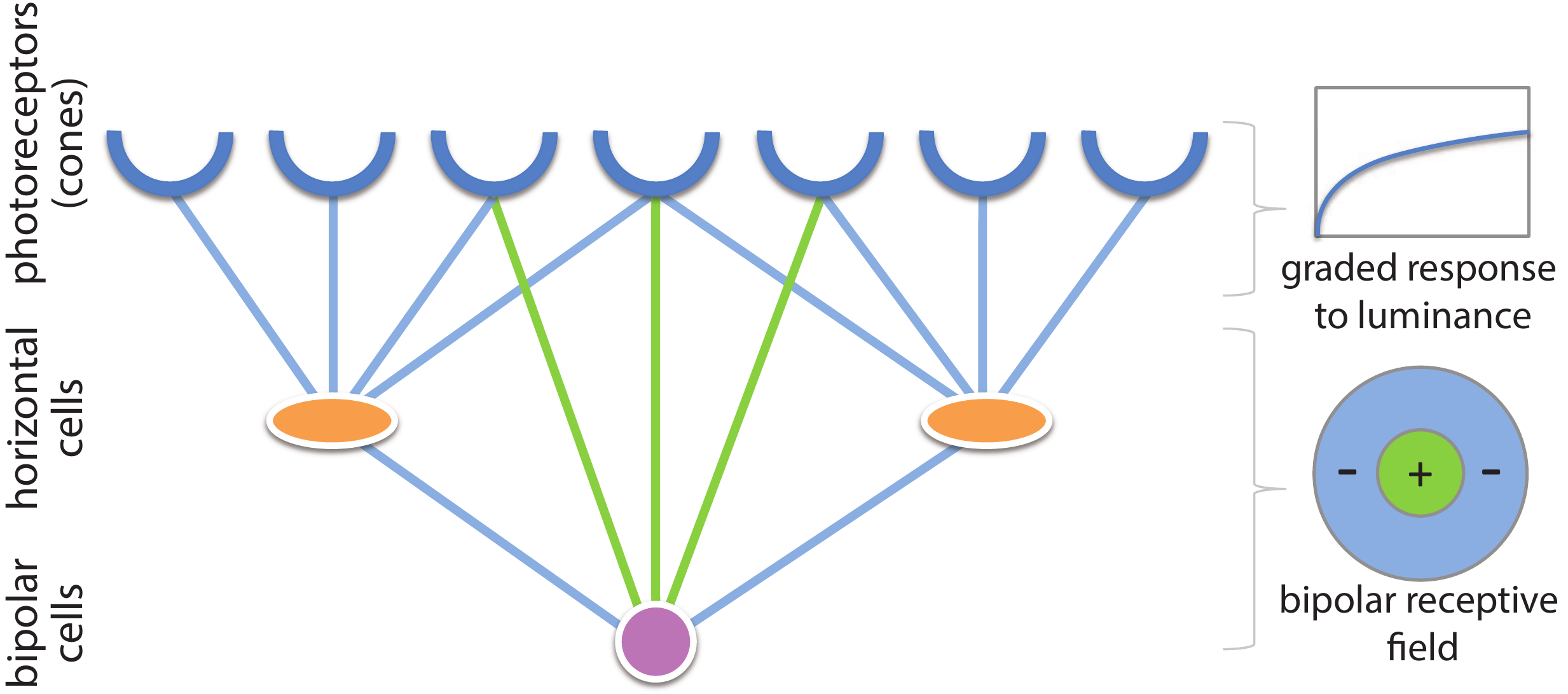}
\caption{\label{fig:retina}Early stages in the human visual system. Top-row illustrates a layer of photoreceptors in the retina. The direct connections (green lines) between the receptors and the bipolar cells form the excitorial center respective field of the bipolar cells. The inhibitorial surround field is mediated by the horizontal cells which are laterally connected (blue lines) to nearby receptors. On the right we depict the graded excitation of the receptors and the center and surround receptive fields of the bipolar cells.}
\vspace{-0.1in}
\end{figure}

\begin{figure}[t]
\centering
\includegraphics[width=2in]{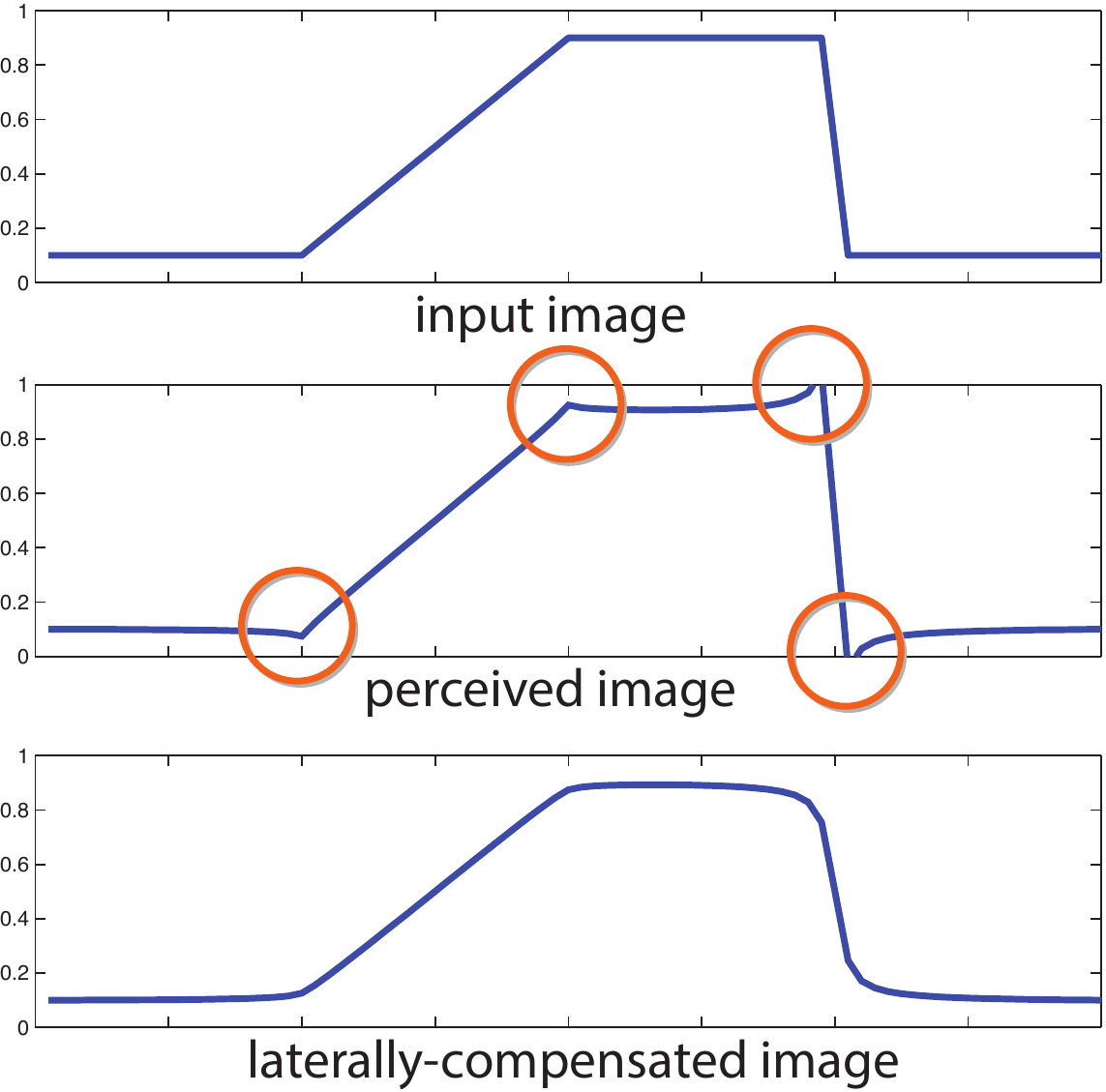}
\caption{\label{fig:examp_1d}Prediction of Mach bands and halos. Top plot shows a scanline taken from an image containing a smooth gradient as well as a step edge. The middle plot shows the corresponding perceived values predicted by the model where the orange circles indicate the formation of Mach bands (left) and halos (right). The scanline at the bottom is taken from the laterally-compensated image that we compute and display.}	
\vspace{-0.235in}
\end{figure}

The Hartline-Ratliff model~\cite{hartline57} describes this mechanism using a two-layered neural network containing inhibitory interactions between nearby neurons. According to this model, given the excitation $e_n$ at the $n$-th receptor (or group of receptors), the response at the corresponding neuron (or group of neurons) in the next layer, $r_n$, is given by
\vspace{-0.1in}
\begin{equation}
\label{eq:HR}
r_n = e_n - \sum_mk_{nm}r_m,
\vspace{-0.1in}
\end{equation}
where $k_{nm} $ are the interaction coefficients of the inhibitory kernel. The interaction strengths are typically assumed to be a function of the distance between neuron, i.e., $k_{nm} = k_{n-m}$ in case of a regular cells arrangement. This model holds for steady viewing conditions in which the system reached a steady-state. Physiological studies suggest that the fixation time period required is a fraction of a second~\cite{biederman71}. Figure~\ref{fig:retina} summarizes the two stages of the visual system we discussed and Figure~\ref{fig:examp_1d} demonstrates its ability to predict Mach bands and halos.



When viewing an image on a display, the image pixels are projected onto the retina and the receptors, depending on their receptive fields, are stimulated by the image pixels and the retinal inhibition process takes place. This process can be expressed in image coordinates by back-projecting the inhibitory kernels onto the grid of the displayed pixels. By transforming Eq.~\eqref{eq:HR} to image coordinates we get the following perceived-image formation model
\begin{equation}
\label{eq:model}
\perc(\x) = \ems(\x) - (\blur * \perc )(\x),
\end{equation}
where $\ems(\x)$ denotes the excitation by the displayed pixel $\x=(x,y)$. The inhibitory kernel $ \blur(\x) $ represents the cones receptive field back-projected onto the display grid. Ghosh et al.~\shortcite{ghosh06} use difference of Gaussians to explain various low-level illusions and Vu et al.~\shortcite{Vu09} use them to achieve invariance to illumination for face recognition. While we explored other types of functions (with convex, concave and linear decaying profiles), we could not establish any conclusive preference by the viewers in a user-study. Thus, we follow existing approaches and use Gaussian functions for defining $\blur(\x)$ by
\begin{equation}
\label{eq:kernel}
\blur(\x) =  \alpha \left(\frac{\exp(-\|\x\|^2/\sigma^2)}{\sqrt{\pi}\sigma}  - \delta(\x)\right).
\end{equation}
While the receptive field is assumed fixed in the retina the kernel scale, $\sigma$, depends on the display's pixel size and, under a perspective viewing model, it must also grow with the vieweing distance. Therefore, we determine $\sigma$ by
\begin{equation}
\label{eq:scale}
\sigma = 7.1 \!\times\!10^{-3} \!\times\! p\! \times\! d,
\end{equation}
where $ d $ is the viewing distance in inches and $ p $ is the display pixel density per inch (assuming the coordinates $\x$ are integer pixel coordinates). The parameter $\alpha$ determines the inhibitory coupling strength or, in our case, the compensation level that we produce. We used $\alpha=0.037$ (along with $\sigma$ defined above) to produce all the results shown in this paper unless noted otherwise. In Section~\ref{sec:userstudy} we explain how we obtained these value and confirmed their optimality by a user-study. In the study we also validated the linear relation between $\sigma$ and the viewing distance $d$.

The second Gaussian in $\blur(\x)$, i.e., the $\delta(\x) $ in Eq.~\eqref{eq:kernel}, describes the sampling unit of the retina. The resolution of current displays is well resolved by the human visual acuity and hence no additional blurring is required. Thus, this Gaussian must be very narrow, effectively a Dirac delta function $ \delta(\x)$, in case of a display with less than 300 pixels per inch. Finally, we note that our choice of  $\blur$, defined in Eq.~\eqref{eq:kernel}, corresponds to a commonly-used filter for image sharpening known as \emph{unsharp-mask}~\cite{Neycenssac93}. Unlike the latter, our compensation scheme consists of inverting an operator containing this term.


\subsection{Producing Laterally-Compensated Images}
\label{sec:contribution}
In order to compute a laterally-compensated stimulus image $\ems'(\x)$ that when viewed on a display no lateral inhibition effects are perceived, we first compute a target image $\perc'(\x)$ describing how an input image is perceived by a visual system that \emph{does not} contain the lateral inhibition mechanism. Given the latter, we recover the laterally-compensated image $\ems'(\x)$ from Eq.~\eqref{eq:model}. The target perceived image $\perc'(\x)$ is obtained by simply omitting the lateral inhibition mechanism from the model. Therefore, by applying only Eq.~\eqref{eq:curve}, we get that $\perc'(\x) = \phi(\lum(\x))$, where $\lum(\x) $ is the intensity of the input image that we wish to display with no lateral inhibition effects.

To summarize, given an input image $ \lum(\x)$, we map it though the response curve $\phi(y) $  to obtain the target perceived image $\perc'(\x)$. We then compute the corresponding stimulus image $ \ems'(\x) $ according to Eq.~\eqref{eq:model} by
\begin{equation}
\label{eq:scheme}
\ems'(\x) = \perc'(\x) +  (\blur * \perc')(\x).
\end{equation}
Finally, we map $\ems'(\x) $ back to image intensity by inverting Eq.~\eqref{eq:curve}, i.e., $\lum'(\x)=\phi(\ems'(\x))^{-1}$, and display the laterally-compensated output image $\lum'(\x)$.

In principle, this lateral-compensation scheme can be carried out using any perceived-image model that accounts for lateral inhibition. Kingdom and Moulden~\shortcite{kingdom88} use a first-order derivative of a Gaussian filter for predicting the appearance of borders (edges). The use of a differentiated Gaussian corresponds to a band-pass filter in which low- and high-frequency components are lost. While this model can be used for predicting perceived quantities, it is impossible to use it in our scheme where the perceived model must be inverted. Moreover, this scheme is limited to one-dimensional signals. As we mentioned above, Vu et al.~\shortcite{Vu09}  use difference of Gaussians for achieving illumination-invariant image representation. However, similarly to Kingdom and Moulden, not all the frequencies exist in this representation and hence this model cannot be inverted.

\begin{figure}[t]
\centering
\includegraphics[width=3in]{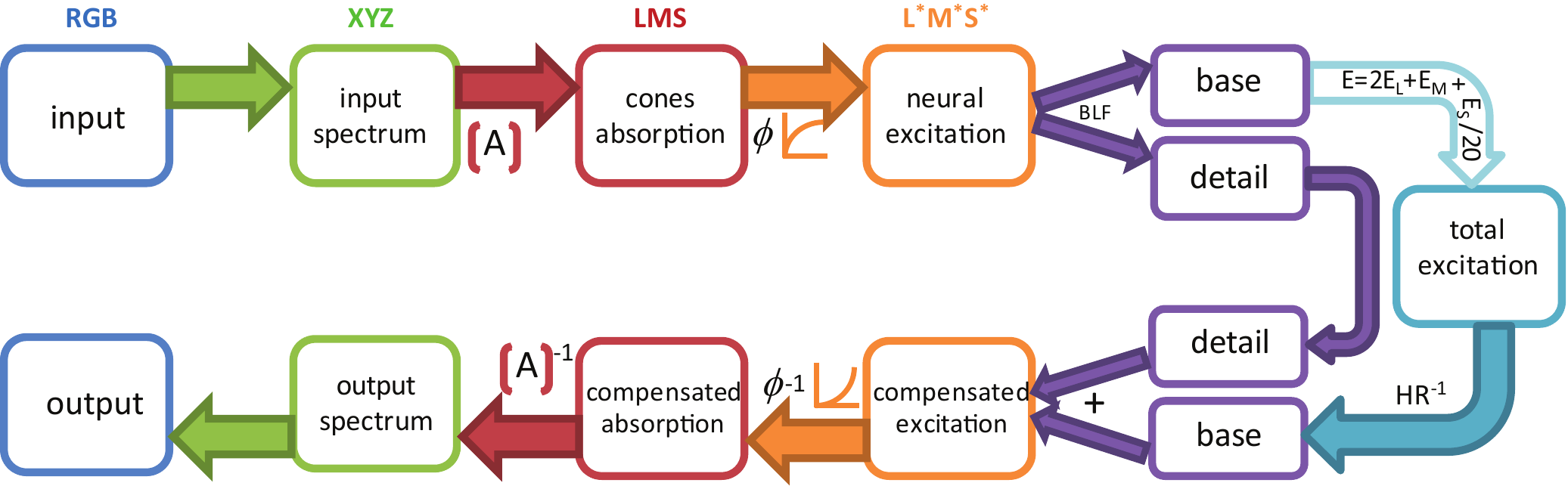}
\caption{\label{fig:system}Lateral inhibition compensation pipeline. Each column shows a different step in the pipeline (top) and its inverse (bottom). The colorspace is noted at the top of each column. The inverse of the Hartline-Ratliff operator (Eq.~\eqref{eq:scheme} or Eq.~\eqref{eq:color_scheme}) is denoted by $\mathrm{HR}^{-1}$. The dashed arrow on the right indicates that no forward inhibition is calculated when producing the target perceived image $\perc'(\x)$. Thus, the output differs from the input only in terms of lateral effects. BLF denotes the optional base-detail decomposition computed by the bilateral filter.}
\vspace{-0.25in}
\end{figure}

\subsection{Color Images}
\label{sec:pipeline}
\vspace{-0.05in}
The procedure described above is applicable to achromatic images. In this section we extend it to conventional RGB images, $\img(\x)$. The mapping between the light absorbed by the cone cells and the neural stimulus they output, in Eq.~\eqref{eq:curve}, is defined over physical light intensity units. Therefore, we need to obtain a description of the light emitted when displaying each RGB pixel value. We achieve this by extracting the display ICC profile and use its associated mapping that approximates the display's light spectrum in the standardized CIEXYZ colorspace. In this colorspace the visible light range is spanned by a linear combination of three non-negative standardized color matching functions $ \bar{x}(\lambda), \bar{y}(\lambda),$ and $\bar{z}(\lambda)$ that specify the emission in each wavelength $\lambda$. Let us denote the spectral sensitivity of the long-, middle- and short-wavelength cone photoreceptors by $ \bar{l}(\lambda), \bar{m}(\lambda),$ and $\bar{s}(\lambda)$ respectively. Given these two triplets of emission and response functions, we compute the total light absorbed by each type of the cone cells being exposed to the light emitted by each of the CIEXYZ color channels by
\begin{equation}
A_{\bar{f} \bar{g}} =  \int \bar{f}(\lambda) \bar{g}(\lambda)  d \lambda,
\end{equation}
where $ \bar{f} $ indexes the cone spectral sensitivity functions, i.e., $ \bar{f} \in\{\bar{l}, \bar{m}, \bar{s} \} $, and $ \bar{g}$ indexes the  CIEXYZ color matching functions, $ \bar{g}\in\{\bar{x}, \bar{y}, \bar{z} \} $. These integrals are evaluated over the visible light wavelength interval $ [380nm, 780nm]$ and result in the following 3-by-3 absorption matrix
\begin{equation}
A=\left[
\begin{array}{ccc}
63  & 74.7 &   7.5 \\
   40.5 &  65.7 &  12.6 \\
   14  &  4.1  & 75.1
\end{array}
\right].
\end{equation}
The product of $A$ with a three-dimensional vector of XYZ color coordinates gives the light absorbed by the long-, middle- and short-wavelength cones, which is a coordinate in the LMS colorspace. Note that unlike the color transformation matrix between the XYZ and LMS colorspaces, the absorption matrix $A$ is not designed to match similar colors between the colorspaces.

Given the CIEXYZ value of each pixel, we use $A$ to obtain the absorbed $L(\x), M(\x) $ and $S(\x)$  values across the entire image. We estimate the graded excitation produced by each cone cell according to the compressive relation in Eq.~\eqref{eq:curve} and use it, as we did above, to define the target perceived image
\begin{equation}
\perc'_L(\x)  = \phi(L(\x)), \: \perc'_M(\x) = \phi(M(\x)), \: \perc'_S(\x) =\phi(S(\x)).
\end{equation}

\begin{figure}[t]
\centering
\includegraphics[width=2.3in]{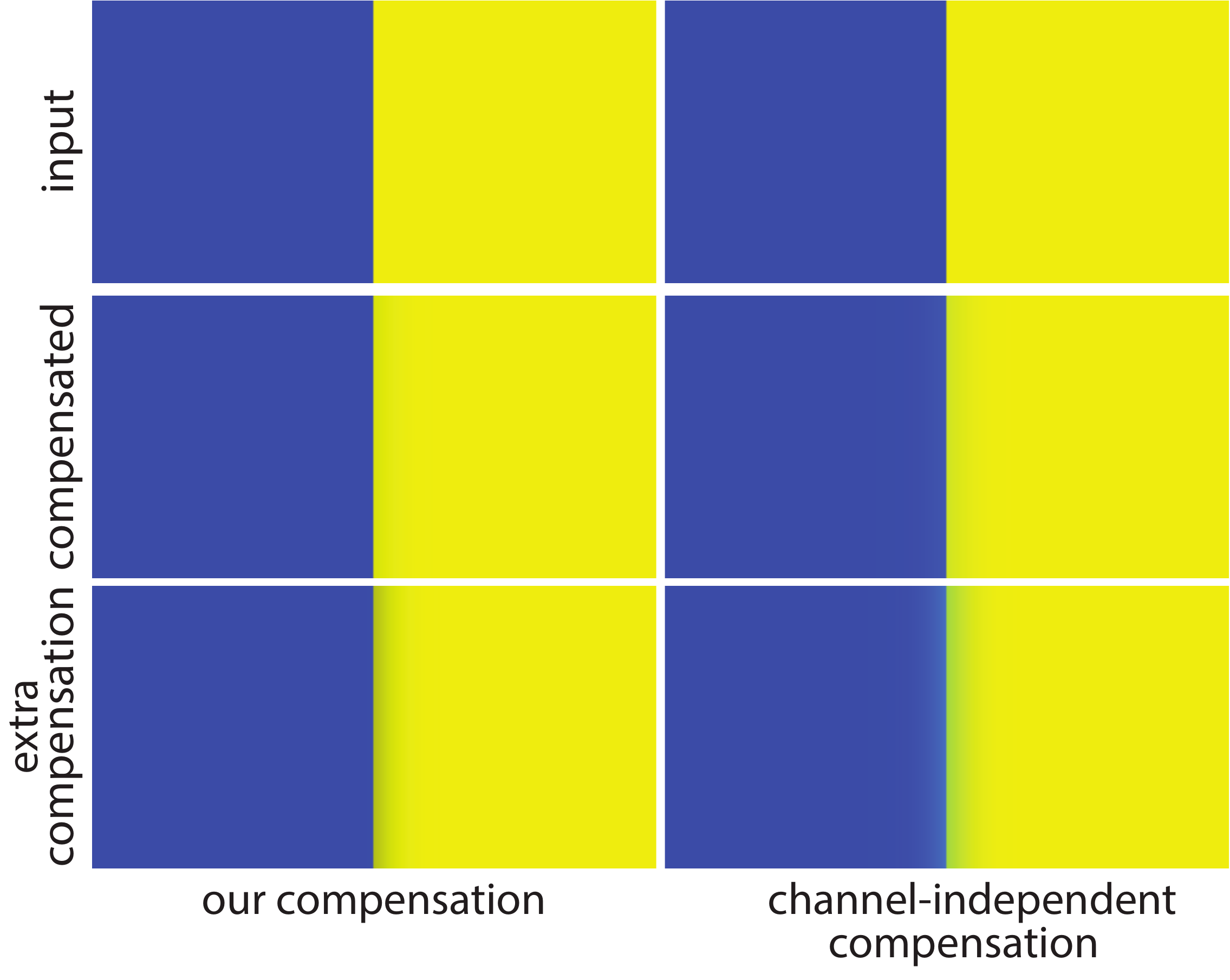}
\caption{\label{fig:chromatic}Chromatic edge between opponent colors. Top row shows the input (two identical images), middle row shows the laterally compensated result by our model (left) and the channel-independent inhibition (right). At the bottom we show an exaggerated compensation ($\alpha=0.25$) by both approaches. In the right column a green halo is observed at the yellow side of the edge.}
\vspace{-0.15in}
\end{figure}

In order to model the lateral inhibition occurring between these three color channels we extend the Hartline-Ratliff relation to account for the three types of cone cells in the human retina. The extension, which we derive at the Appendix Section, found in the supplemental material, is based on the physiological observation that the connections between the horizontal cells and the cone cells are chromatically blind, i.e., they do not exhibit a chromatic preference~\cite{boycott87,dacey96}. Thus, in our derivation we assume that the long- , medium-, and short-wavelength cones equally inhibit one another and obtain an inhibition model, analog to Eq.~\eqref{eq:scheme}, given by
\begin{equation}
\label{eq:color_scheme}
\ems'_{L/M/S}(\x) = \perc'_{L/M/S}(\x) +  (\blur * \perc' )(\x),
\end{equation}
where subscripts $L/M/S $ denote three separate equations in the LMS space. The common term  $\perc'(\x)$ describes the total excitation in the retina resulting from all three cell types. As we explain in the Appendix, since there are typically 50\% more long-wavelength cones than medium-wavelength cells~\cite{carroll02} and the short-wavelength cones constitute about ten percent of the total cones population in the fovea, the total excitation is computed by
\begin{equation}
\perc'(\x) = \sfrac{3}{2}\perc'_L(\x) + \perc'_M(\x) + \perc'_S(\x)/4.
\end{equation}
A similar quantity is often used to estimate the achromatic component of images~\cite{martin98,pattanaik00}. Note that if we replace the total excitation in Eq.~\eqref{eq:color_scheme} with the excitation in each color channel, i.e., use $\blur * \perc'_{L/M/S} $ to describe the lateral inhibition, chromatic effects attempting to correct chromatic inhibition biases will appear. Whether chromatic lateral inhibition effects occur in the human retina is a matter of debate. Reports that describe chromatic biases note that their behavior differs from the achromatic lateral inhibition effects~\cite{Spitzer09} and the former is typically associated with successive-contrasts (afterimages) in case of prolonged viewing followed by eye movement~\cite{ware87}. Figure~\ref{fig:chromatic} shows an example containing a strong edge between two opponent colors. In this example we compare the results obtained by our chromatically-blind scheme and the channel-independent alternative, i.e., the use of $\blur * \perc'_{L/M/S} $ in Eq.~\eqref{eq:color_scheme}, which assumes inhibition occurs only between cells of the same type. While our result does not introduce new colors to the compensated image, a thin green halo appears in the channel-independent compensation.

Finally, we obtain the laterally-compensated image by reversing the model transformations in the pipeline we described, i.e., we recover the corrected LMS values by computing $\phi^{-1}(\ems'_{L/M/S}(\x))$, we map these values to  the corresponding CIEXYZ stimulus values using $A^{-1}$ and finally transform back to the device RGB values based on the display's ICC profile. Figure~\ref{fig:system} summarizes the steps in this computational pipeline.

\vspace{-0.03in}

\subsection{Detail Preservation Step}
\label{sec:details}
\vspace{-0.03in}
Since the lateral inhibition mechanism acts as a sharpening operator, its inversion acts as a blurring mechanism. Thus, our compensation procedure acts as a blurring operator which may undermine the contrast of fine image details. While there are many types of images that benefit from our correction, as shown in Section~\ref{sec:results}, there are classes of images that contain important fine details, such as the case of natural images. We circumvent this visual degradation by excluding the details from the compensation procedure. This is carried out by separating the perceived image $\perc'_{L/M/S}(\x)$ into a base and detail components using the bilateral filter~\cite{Tomasi98} (we used $\sigma_s\!=\!5$ and $\sigma_r\!=\!0.08$). The strong edges which typically produce the lateral-inhibition halo effects are left in the filtered base image, $B_{L/M/S}(\x)$, and hence we apply our correction only to this component of the image. Finally, we add the original detail component of the image, $D_{L/M/S}(\x)\! =\!\perc'_{L/M/S}(\x)\!-\!B_{L/M/S}(\x)$, to the resulting compensated image $\ems'_{L/M/S}(\x) $.


\vspace{-0.03in}

\begin{figure}[t]
\centering
\includegraphics[width=2.8in]{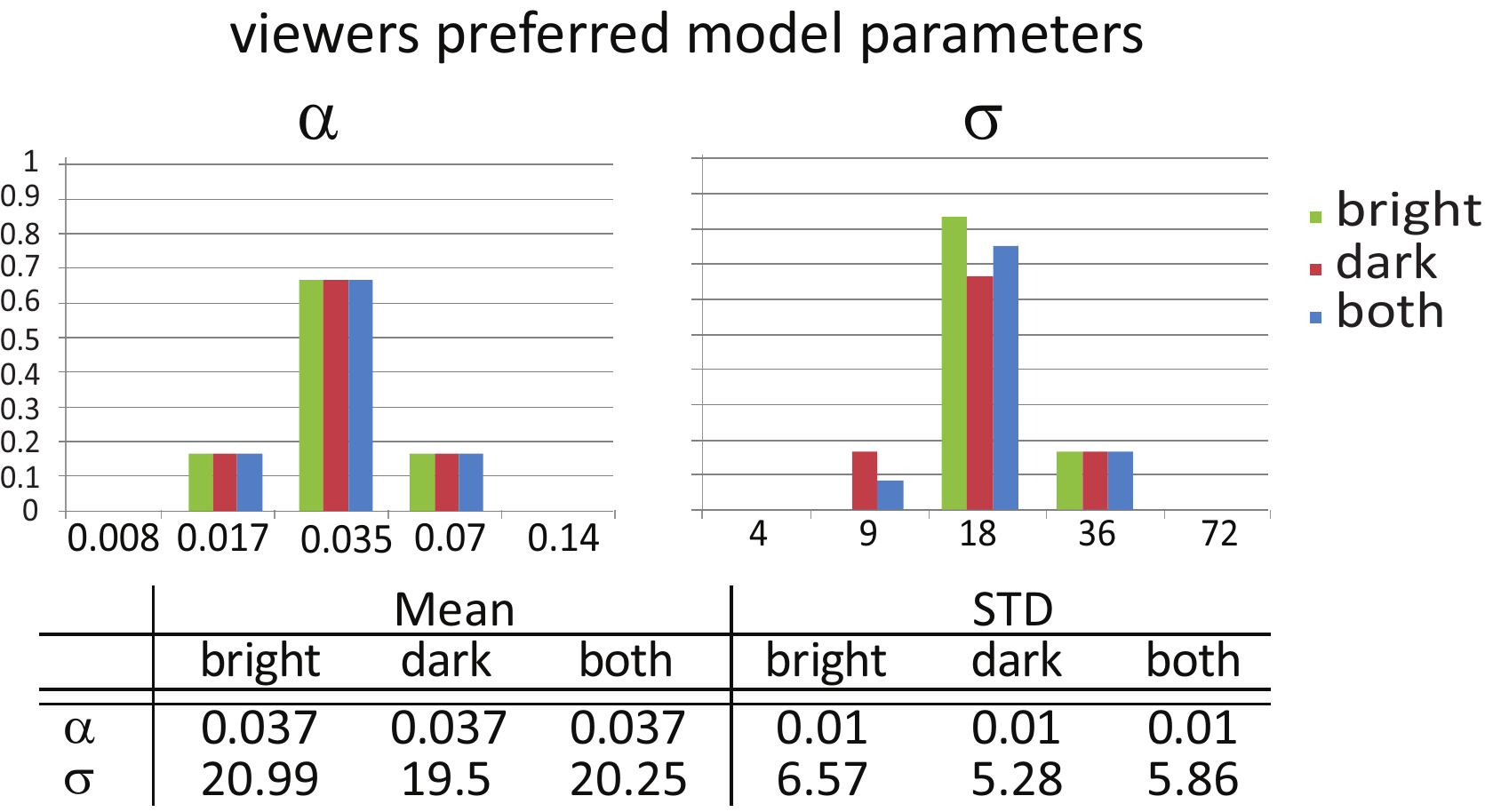}
\caption{Compensation level and scale parameter evaluation. Bars show the proportion of viewers who preferred each value of $\alpha$ (left) and of $\sigma$ (right). Bar colors indicate display brightness setting.}	
\label{fig:slider}
\vspace{-0.15in}
\end{figure}


\section{Evaluation and Results}
\label{sec:results}

\vspace{-0.01in}
In order to implement and evaluate our method we used screen calibration equipment (X-Rite Eye-One Display) to extract the ICC profile of the display we use. This provides us the mapping between raw RGB image values and the radiometrically-linear CIEXYZ coordinates parametrizing the spectrum of the light emitted from the display. In order to avoid banding effects in the corrected images we produce, we conducted our evaluation on a 30-bit display (Dell U2410). Such displays are becoming affordable and popular. Our method mostly consists of pointwise operations (colorspace transformations) and a single convolution with a fairly small kernel. Our Matlab implementation takes 0.75 seconds to process a one-megapixel image on a 2.6Ghz Intel Core i7 machine.

\vspace{-0.01in}
While the reader cannot recreate the exact viewing conditions for which we computed the compensated images, we would like to encourage him or her to view our images in the supplemental material, where they are more genuinely reproduced than in the paper. In order to find an effective viewing position we advise testing different viewing distances, around 30 inches between the forehead and the display in case of a 24 inch display. This distance grows linearly with the display size.
\vspace{-0.05in}

\subsection{User Study}
\label{sec:userstudy}
\vspace{-0.05in}
We conducted a user study consisting of 22 volunteers with several goals in mind; calibrating our model by estimating its parameters, comparing its accuracy to alternative inhibition models, evaluating its effectiveness in reducing biases due to lateral inhibition and studying the method's dependence on viewing conditions. The room in which we conducted the tests was slightly dimmed. When seating the participants we made sure their forehead is approximately 30 inches away from the display. We started each experiment by ensuring the participant understands the phenomenon we are referring to and intend to cancel. For example, we verified that the viewers observe the lack of uniformity in certain regions and halos around particular edges. The participants were selected from different disciplinary backgrounds and were naive as to the purpose of the study. They had normal or corrected-to-normal vision. Six of the viewers participated in the model calibration described below and the rest (sixteen) for evaluating our method in a sequence of tests described in this section.


\begin{figure}[t]
\centering
\includegraphics[width=2.5in]{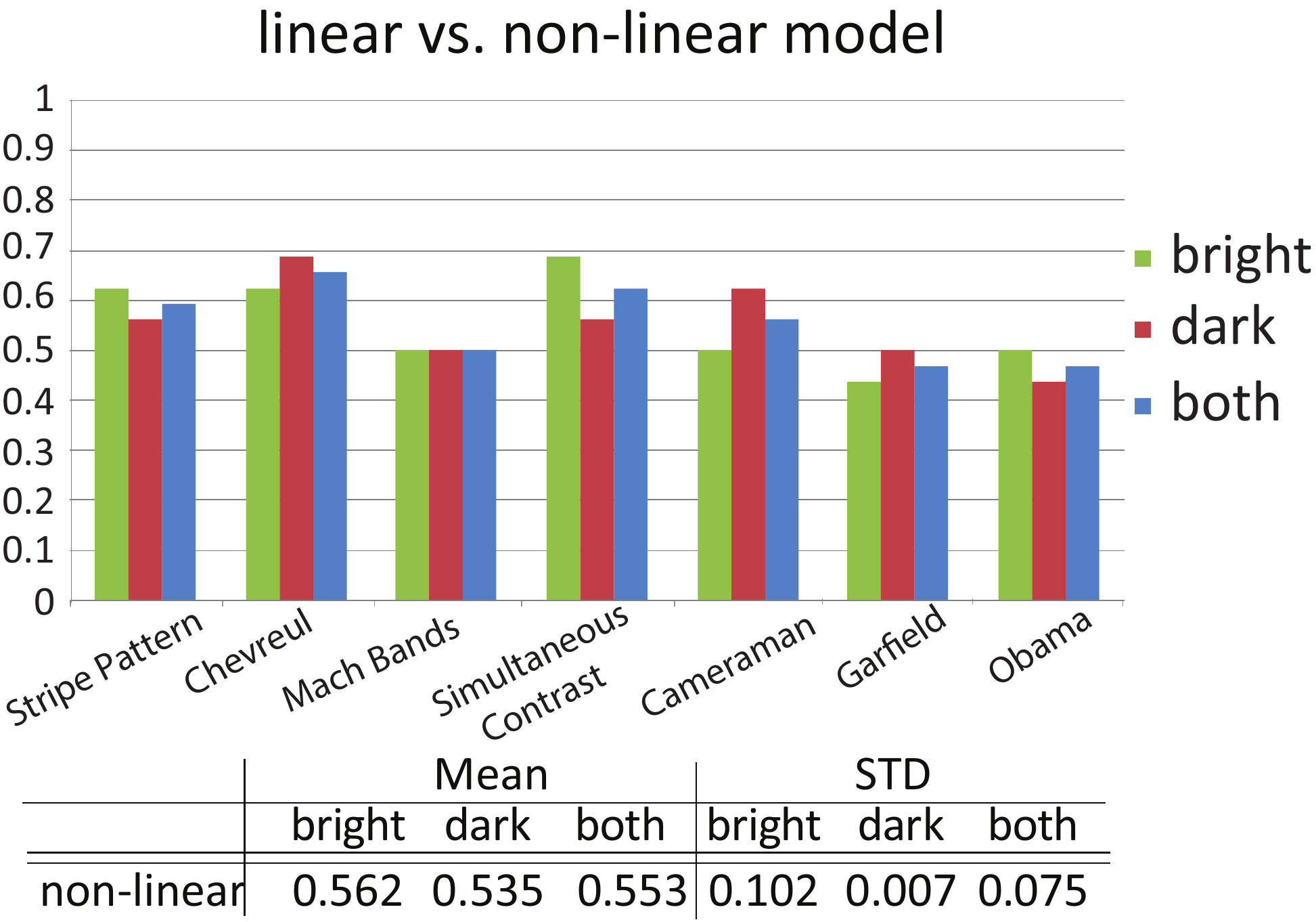}
\caption{Inhibition Model Evaluation. The bars show the number of viewers who found the lateral compensation using Hartline-Ratliff formula to be more effective than Barlow and Lange model. The results do not indicate a strong preference to either model in case of brightness levels produced by standard displays. Bar colors indicate display brightness setting.}	
\label{fig:non_linear}
\vspace{-0.25in}
\end{figure}

\vspace{-0.01in}
\bt{Model Fitting.} The first test is designed to estimate the parameters of the perceived image model, in Eq.~\eqref{eq:model} and Eq.~\eqref{eq:kernel}, namely the compensation level $\alpha$ and the inhibitory kernel base scale in Eq.~\eqref{eq:scale}, that achieve the most effective results. In this test we allowed the participants to adjust a slider that changes $\alpha$ in a discrete range and displayed the resulting laterally-compensated image interactively. The participants were asked to search for the configuration in which the effects due to lateral inhibition are minimized. In this task we used the Stripes Pattern (shown in Figure~\ref{fig:teaser}) and asked the viewers to achieve the most uniform gray stripe possible. The same step was conducted in order to determine an optimal value for $\sigma$ which determines the base scale according to Eq.~\eqref{eq:scale}. We repeated this process by alternating between the two parameters until the changes were insignificant. The lateral-inhibition effects and their reduction are most-noticeable on the Stripes Pattern and hence we conducted the fitting using this image.

\vspace{-0.01in}
Figure~\ref{fig:slider} shows the histogram of the values selected by the viewers at the last iteration when we set $\alpha=0.035$ and $\sigma=20$ (at particular viewing conditions specified below). In the case of the compensation level $\alpha$, more than 65\% of the participants choose the value $\alpha = 0.035$ and no other value received more than 17\% of the votes. The estimated mean in this case is $\alpha=0.037$ with STD of $0.01$. In case of $\sigma$ almost 70\% of the participants voted for $\sigma=18$ and the computed mean is $20.25$ with STD 5.86. In this test the viewing distance is $d=30$ inches and display pixel density of $p=94$ pixels per inch. Thus, according to Eq.~\eqref{eq:scale}, the value of $\sigma=20.25$ translates into the base kernel value of $7.1 \!\times\!10^{-3} $. We used the average values of $\alpha $ and base scale obtained in the last iteration to produce all the results in the paper unless stated otherwise.


\begin{figure}[t]
\centering
\includegraphics[width=2.7in]{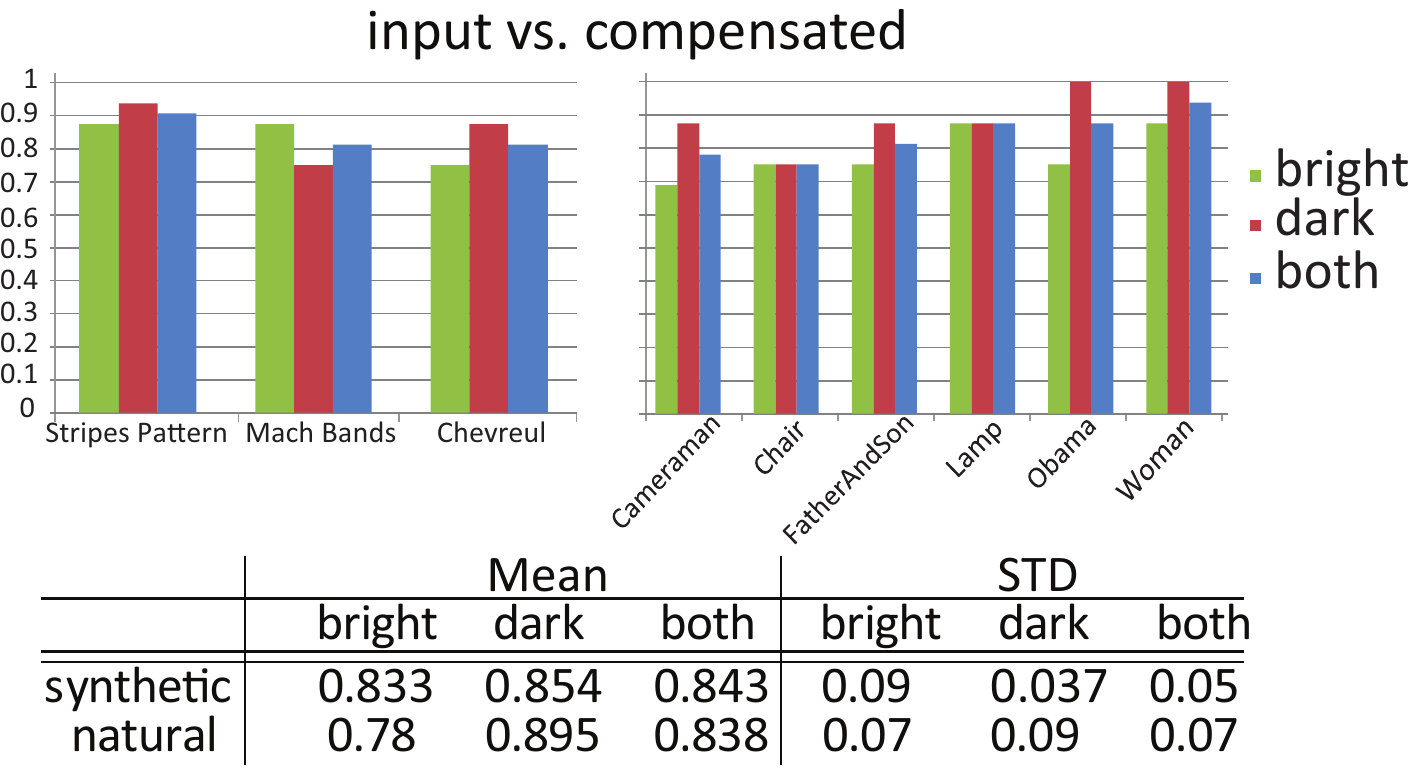}
\caption{Effectiveness in reducing lateral biases. Graphs show the proportion of viewers who indicated that the compensated images show reduced lateral inhibition effects compared to the input. Left graph shows the results obtained from the synthetic patterns and right graph shows the results over natural images. Bar colors indicate display brightness setting.}	
\label{fig:input_output}
\vspace{-0.2in}
\end{figure}


\vspace{-0.01in}
\bt{Model Accuracy.} As noted in Section~\ref{sec:method}, Barlow and Lange~\shortcite{Barlow74} suggest a model in which the amount of inhibition depends on the excitation level. In this model the inhibition term $(\blur * \perc )(\x)$ in Eq.~\eqref{eq:model} is multiplied by $(1-\beta \ems(\x))$ where $\beta > 0$. We conducted a user-study to evaluate the accuracy of the perceived image model derived in Section~\ref{sec:method} versus the option of adding this dependency on the excitation level. In this test we displayed the images corrected using each of the models side-by-side (at a random order) and asked the viewers to pick the image containing the least amount of inhibition effects. Figure~\ref{fig:non_linear} shows the viewers choices which is centered around 55\% (in favor of the original Hartline-Ratliff based compensation) with STD of 7.5\%. This trend is consistent in both display brightness settings which indicates that the dependence between the levels of the inhibition and excitation is negligible at the brightness regimes of standard displays.

\vspace{-0.01in}
Since the model of Barlow and Lange~\shortcite{Barlow74} does not show improvement despite its higher computational cost, we focused on the perception model defined in Eq.~\eqref{eq:model} and conducted all the rest of the tests reported in this paper using it. In this test we determined the optimal value of $\beta$ (and its matching $\alpha$ and $\sigma$ values) using the same procedure described above for the Hartline-Ratliff model.

\vspace{-0.01in}
\bt{Method Effectiveness.} To measure the effectiveness of our approach we presented to the participants the input image next to our output image (again, at a random order) and asked them to specify where the biases due to lateral inhibition appear weaker. Figure~\ref{fig:input_output} shows that 84\% of the viewers picked our result in case of the synthetic patterns and 83\% in case of the natural images. Below we report various other, more specific, tests whose success reflects positively on the effectiveness of our approach.

\begin{figure}
\centering
\includegraphics[width=2.3in]{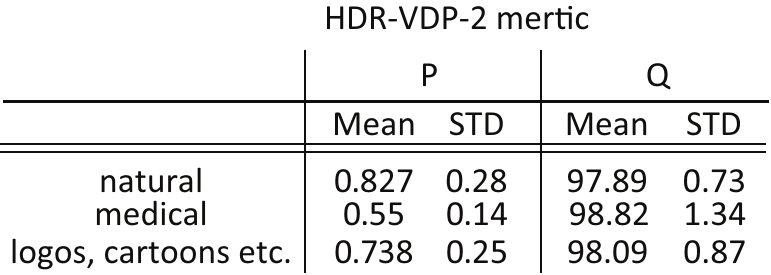}
\caption{Scores obtained from the HDR-VDP-2 metric. The results are divided into three classes of images: natural images, medical images (X-ray) and logos, maps, cartoons, technical drawings images. The P value predicts the probability that the compensation is visible to an average observer and the Q value predicts the mean-opinion-score of the quality degradation by our method.}
\label{fig:vdp}
\vspace{-0.18in}
\end{figure}

\vspace{-0.01in}
As another test for estimating the overall performance of the method, we evaluated it using the visual metric of Mantiuk et al.~\cite{Mantiuk11}. This metric is expected to predict the likelihood at which an average viewer will notice the difference between the input and the resulting corrected images as well as the quality degradation associated in the process. The table shown in Figure~\ref{fig:vdp} provides the two scores predicted for three different classes of images. According to this metric, the chances at which the differences will be visible are 82.7\% in case of natural images, 55\% in case of medical, and 73.8\% in case of maps, cartoons and technical drawings. While only every other viewer is predicted to notice our lateral-compensation in case of medical images, we believe that this figure will increase in case of trained radiologists. We should also note that some of the other tests reported here indicate higher success rates. As for the quality degradation, in all three image categories, our method obtained a mean-opinion-score of 97\% or higher.

\begin{figure}[t]
\centering
\includegraphics[width=1.7in]{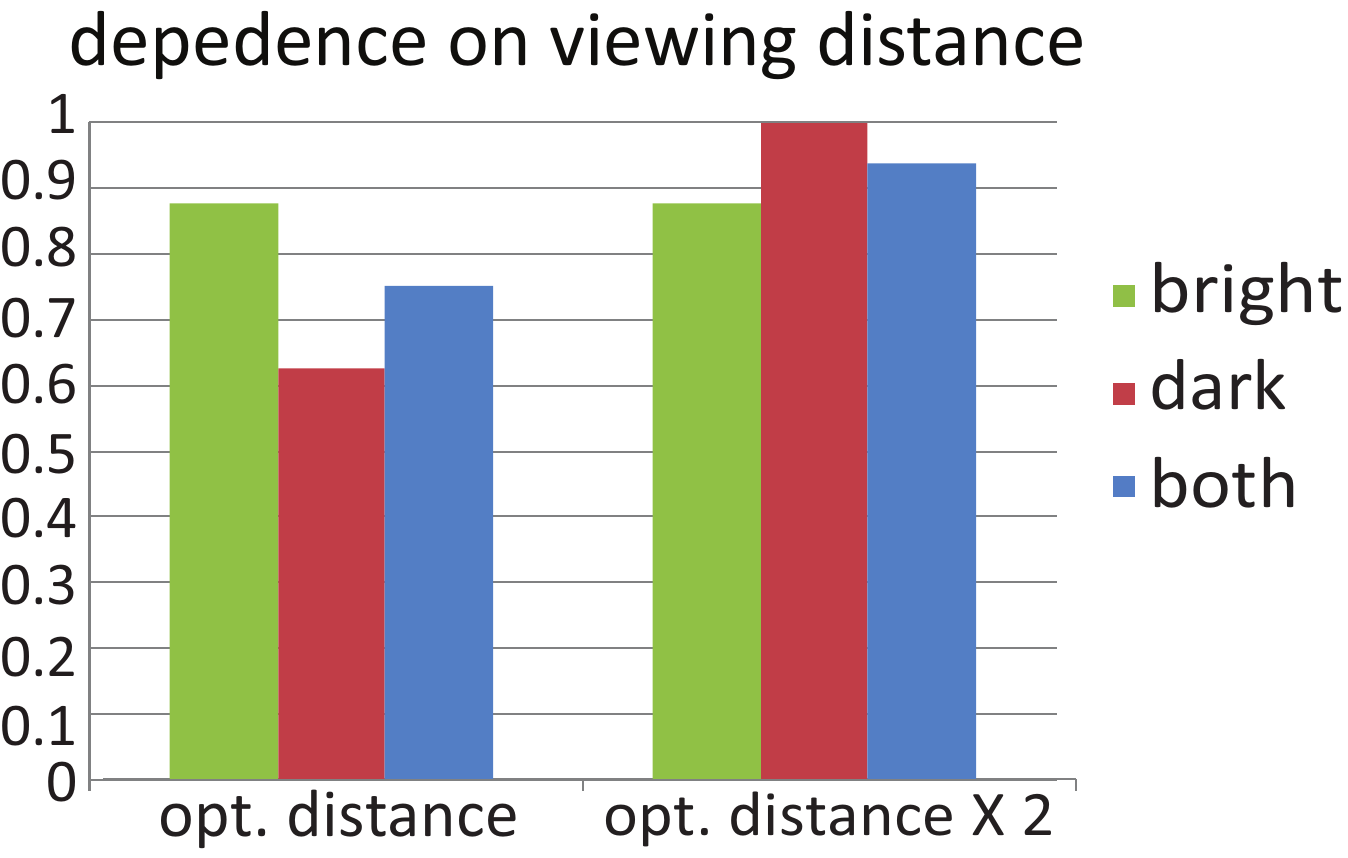}
\caption{Viewing distance. 75\% of the viewers picked the smaller kernel when sitting close to the display and almost all the viewers chose the larger kernel when seated farther. Bar colors indicate display brightness setting.}	
\label{fig:distance}
\vspace{-0.39in}
\end{figure}

\begin{figure}[t]
\centering
\includegraphics[width=2in]{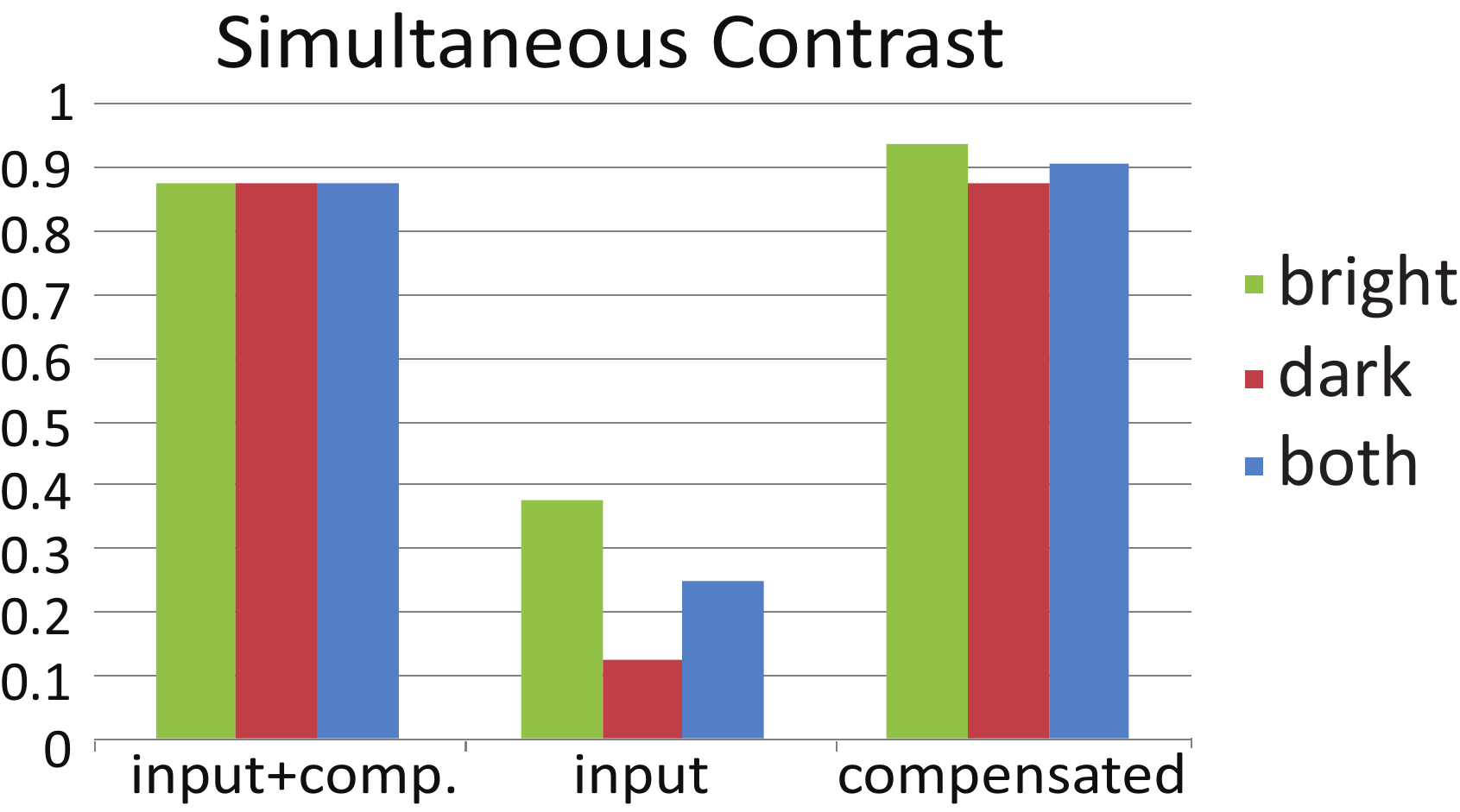}
\caption{Simultaneous Contrast Test. The left bars show the proportion of viewers who found circles' brightness closer in our compensated image versus the input image (when watching both images at the same time). The center and right bars show the proportion of viewers who indicated that the circles have an identical brightness in the input image (center bars) and in our compensated image (right bars). Bar colors indicate display brightness setting.}
\label{fig:sim_con_graph}
\vspace{-0.2in}
\end{figure}

\vspace{-0.01in}
\bt{Display Brightness.} We also wanted to check at what extent the compensation we compute depends on the display brightness setting. To do so, we conducted all the tests described in this section under two display settings; a bright mode in which the maximal intensity is 240 $\tfrac{cd}{m^2}$ and a dim mode with 50\% this intensity. The results shown in figures~\ref{fig:slider},~\ref{fig:non_linear},~\ref{fig:input_output},~\ref{fig:distance},~\ref{fig:sim_con_graph} and~\ref{fig:medical} by the green- and red-color bars do not indicate that there is a major dependence on the display brightness. For example, in Figure~\ref{fig:slider} there is a difference of less than 6\% in the estimated $\sigma$ and no difference in $\alpha$, and in Figure~\ref{fig:input_output} there is less than 3\% difference in the viewers choice in case of synthetic patterns and 13\% in case of the natural images. Moreover, these differences do not correspond to a consistent trend in which the method operates better or worse and hence we conclude that it is invariant to moderate changes in the display brightness level.


\vspace{-0.01in}
\bt{Viewing Distance.} In another test we displayed, side-by-side (at a random order), two laterally-compensated images produced using the optimal kernel scale $ \sigma $ found and using twice that value. The participants were seated at two different distances from the display; at the distance we conducted the previous test (30 inches between the forehead and display), and optimized the kernel scale for, and twice that distance (60 inches). In both cases the participants were requested to pick the image in which the horizontal gray stripe appears more uniform in the Stripes Pattern image. Figure~\ref{fig:distance} shows that 75\% of the viewers chose the smaller kernel when sitting close to the display and almost all the viewers chose the larger kernel when seated farther. These results support the correction of the kernel scale $\sigma$ based on the viewing distance according to Eq.~\eqref{eq:scale} which assumes perspective viewing model. We believe that this relation should be sufficiently accurate for the range of distances displays are typically viewed.

\vspace{-0.01in}
\bt{Brightness Constancy.} In order to evaluate whether our approach improves the detection of constancy in brightness we processed the Simultaneous Contrast image, found at the supplemental material, and used it to conduct two tests. In the first test we presented the viewers with the input and the compensated images one after the other at a random order. In each case we asked them whether the circles appear to have an identical brightness or not. In the second test we displayed both images one next to the other, again at a random order, and asked the participants to choose the pair in which the circles appear to have closer brightness. Figure~\ref{fig:sim_con_graph} shows that more than 90\% of the viewers indicated identical brightness when viewing our compensated image where as only 25\% of the viewers indicated this is the case when viewing the original image. Similarly, when being presented with both images 87\% of the viewers picked our compensated image as the one in which the circles' brightness appear closer.

\vspace{-0.01in}
We proceed by demonstrating the effectiveness of our lateral compensation for various applications and image types.



\subsection{Applications}
\begin{figure}[t]
\centering
\includegraphics[width=2.5in]{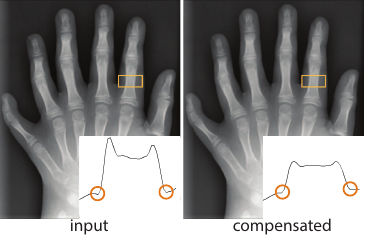}
\caption{X-ray test images. Left image shows the input and right the compensated output. In both images we highlighted the same small horizontal segment (inside the orange rectangle). The 1D plots show the perceived values of the two images along the highlighted segments. Notice the under-shoots between the soft tissue (inside orange circles) which appear in the perceived input and their absence in the compensated output. In the actual test we placed the images and the 1D functions such that their association is not revealed (unlike what is shown here).}	
\label{fig:medical_examp}
\vspace{-0.33in}
\end{figure}

\vspace{-0.05in}
\bt{Medical Data Visualization.} In the next test we wanted to check whether our method improves the accuracy at which visualized data is perceived and used X-ray images for this purpose. To carry this out we computed the perceived values of the input and compensated output along a short horizontal segment, containing an interface between a soft tissue and a bone, to obtain two 1D functions. Figure~\ref{fig:medical_examp} shows the input and output images as well as their corresponding 1D perceived functions. The perceived input values, $\perc(\x)$ along the segment, were computed by solving Eq.~\eqref{eq:model} given $\ems(\x) =\phi(\lum(\x))$. The perceived output values, $\perc'(\x)$ along the segment, are by construction given by $\phi(\lum(\x))$. Given this data (two input images and their two corresponding 1D functions) we conducted the following three tests.


\vspace{-0.01in}
In the first test we displayed the input image with the short horizontal segment highlighted. We plotted the two perceived 1D functions next to the input image and asked the viewers to pick the function which they think best represents the intensity of the highlighted image segment. In the second test we replaced the input image with our compensated output image, kept the same two plots, and asked the same question as in the first test. In the third test we displayed both images and both plots (at a random arrangement) and asked the viewers to match each plot with a different image. These three tests were presented in a random order to the participants.

\begin{figure}
\centering
\includegraphics[width=3in]{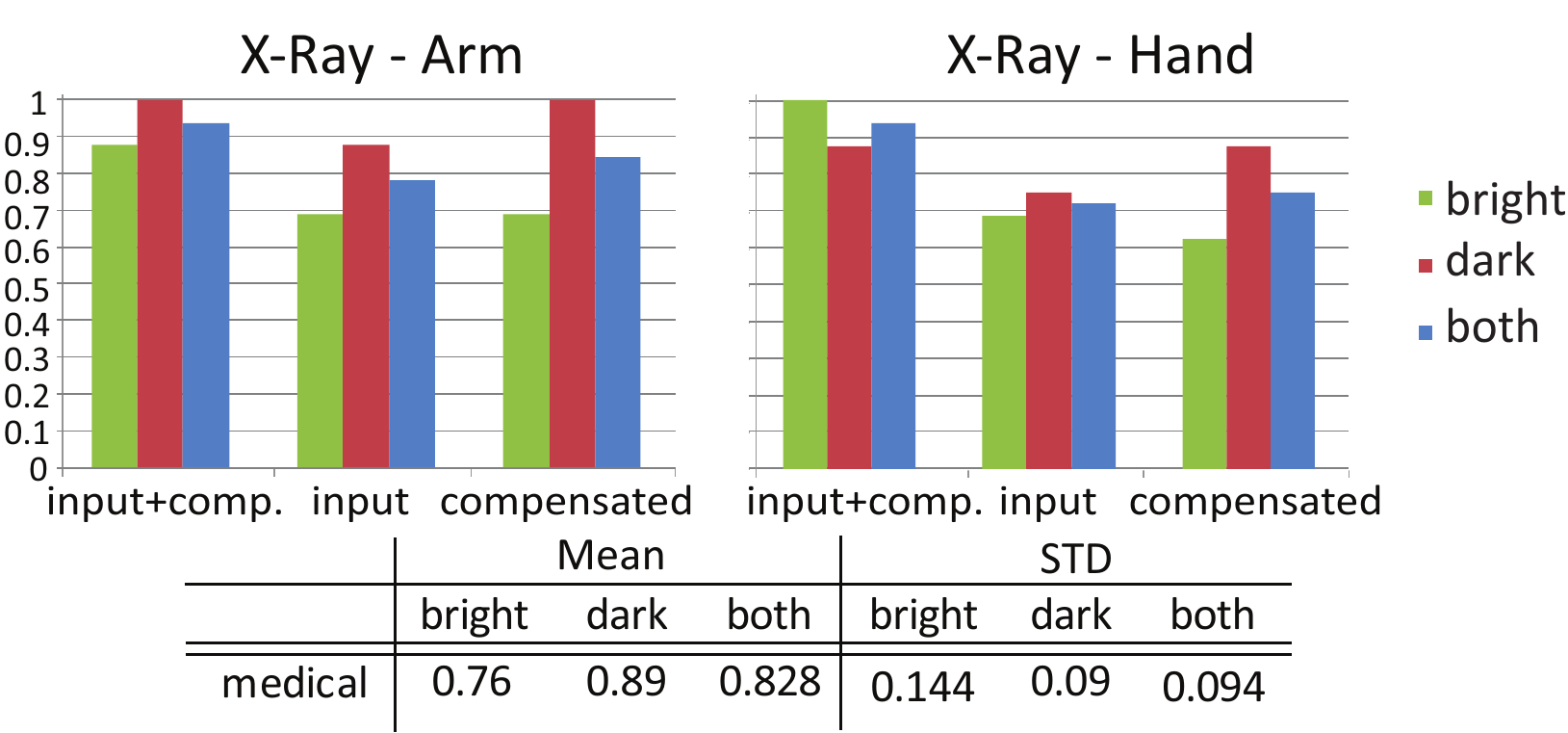}
\caption{Perception of X-ray data. Graphs show the proportions of correct matches between the images and the 1D perceived functions. The two graphs correspond to two different images that we used in this test. The labels at the bottom indicate the image shown to the participant (the three options correspond to the three tests we described). The graphs show that 80\%  of the viewers selections are consistent with our model prediction. Bar colors indicate display brightness setting.}
\label{fig:medical}
\vspace{-0.32in}
\end{figure}

\vspace{-0.01in}
Figure~\ref{fig:medical} shows that in 80\% of the cases, the viewers' choice is in agreement with our model's prediction. This implies that our compensated image conveys the corrected plot that \emph{does not} contain the lateral inhibition effect, the spurious undershoot in the soft tissue near the bone (highlighted by the orange circle in Figure~\ref{fig:medical_examp}), and it is therefore more faithful to the original data which does not contain this feature. The second X-ray referred to in the figure is shown in the supplemental material of the paper.

\vspace{-0.01in}
While these results demonstrate improvement in perceiving the visualized data, radiologists are aware of these biases and perform their diagnosis by taking them into account (despite the concerns raised in~\cite{daff77,nielsen01}). Nevertheless, we approached two senior radiologists and asked them to evaluate the results of our method on six X-ray images, X-Ray 1 up to 6 in the supplemental material taken from~\cite{nielsen01}.  We further asked for their opinion on the plausibility of using such a correction in the diagnosis process. In particular, we first made sure that they understand what our correction does over synthetic patterns (and indeed they were both familiar with the issue under the general term \emph{mach-effects}). Then we showed them pairs of images consisting of input and our output and pointed them to the particular clinical uncertainly discussed by~\cite{nielsen01} in each image. We asked them whether they believe the illusions are reduced and asked for their overall opinion of our output. The head of the radiology dept. recognized the biases in X-Ray 1 up to X-Ray 5 and confirmed that indeed they were weaker in our images but could not identify the artifact in X-Ray 6 and hence its reduction. While he was very positive about integrating such a solution, he raised concerns about the overall loss of contrast in the image. As a compromise he suggested that our lateral correction will be used selectively across the image, where the radiologists will be able to mark a window in which it will apply as well as be able to switch the correction on and off, similarly to other mappings radiologists use when inspecting X-ray images. We received a similar feedback from the oral X-ray specialist we interviewed. She appeared to be very aware of these issues and showed us additional cases where these biases appear. While she was able to see the reduction of the effects in X-Ray 1 to X-Ray 6, she appeared to be more concerned as to the loss of fine details. She, too, suggested that our solution will be used with the ability to toggle between the input and output images.

\vspace{-0.01in}
We believe this feedback confirms that our correction has the potential of aiding medical imaging however it cannot be put to immediate use. A more thorough clinical experimentation is required in order to understand how the corrected images should be incorporated in the diagnosis process.

\vspace{-0.01in}
\bt{Lateral Inhibition Study.} There are several established test patterns that are used for studying lateral inhibition phenomena. We ran our method on several commonly-used patterns and observe a noticeable reduction in the associated effect. As shown in Figure~\ref{fig:input_output}, 84\% of the viewers saw reduction in our corrected images. More specifically, in the Chevreul illusion, found in the supplemental material, false gradients are perceived at the interface between pairs of constant-gray columns. In our output image the columns appear more uniform and less affected by their neighboring columns. A similar reduction of false variations in brightness is observed at the gray horizontal stripe in the Stripes Pattern shown in Figure~\ref{fig:teaser}. A canonical test pattern showing the formation of Mach bands around a smooth gradient and the reduction of these bands in the corrected image can be found at the supplemental material.

\vspace{-0.01in}
Our method's ability to reduce these effects in the test patterns supports the theoretical association between these illusions and the lateral inhibition mechanism. Moreover, our bias-compensation methodology can be used as a framework for studying this mechanism and evaluating different models that describe it.

\vspace{-0.01in}
\bt{Logos, Maps, Cartoons and Technical Media.} Beside visualizing data accurately demonstrated in the previous section, our method is useful for different types of images used in the media and for design. Generally speaking, images that contain large regions of a constant color separated by sharp transitions are prone to halo effects due to lateral inhibition. These emphasized transitions may appear too harsh and the loss of uniformity near these edges may be misleading as to the content of the image.

\begin{figure}[t]
\centering
\includegraphics[width=3in]{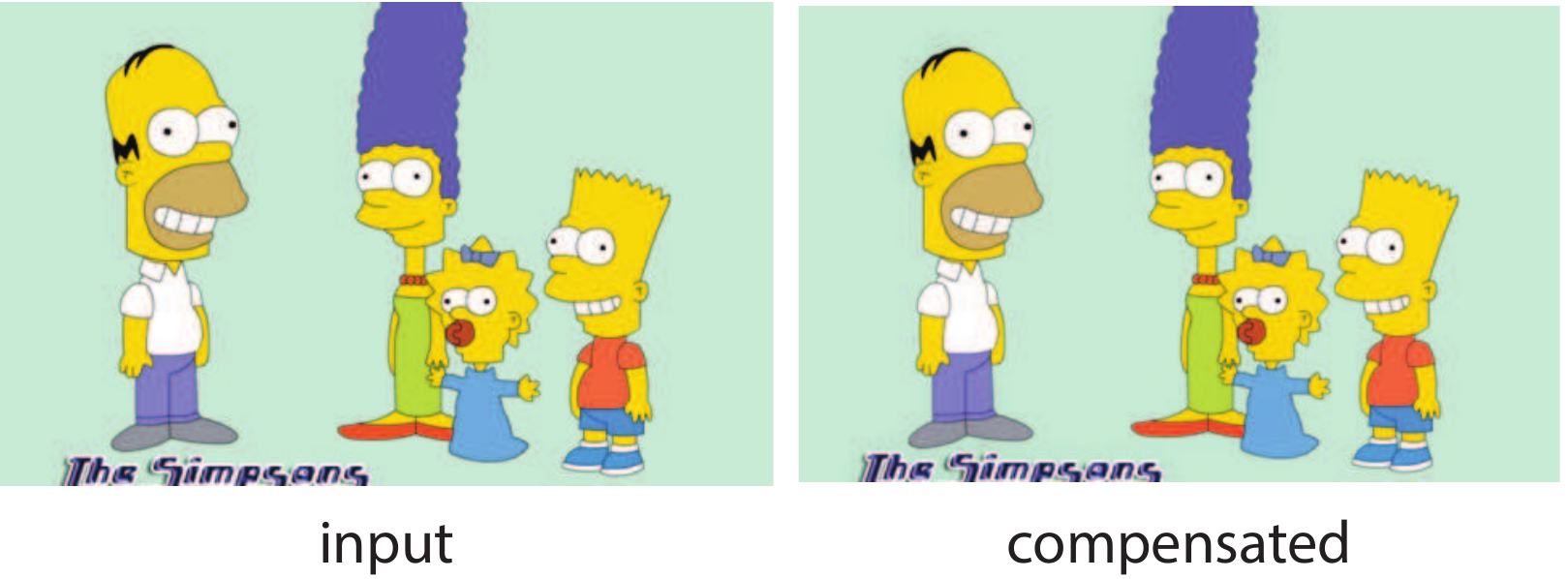}
\caption{Cartoon Drawings. Over-emphasized transitions as well as halos are observed around the characters' outline (mostly around Marge's hair). These effects are reduced in our laterally-compensated image which contains softer edges and a more uniform background.}

\label{fig:cartoons}
\vspace{-0.25in}
\end{figure}

\vspace{-0.01in}
Figures~\ref{fig:teaser} and~\ref{fig:cartoons} show cartoon drawings where halos are observed around the characters' outline. Our method reduces the halos and makes the background appear more uniform. This result may be closer to the artist's intent and/or aesthetically preferable. In the supplemental material we also include cartoon videos processed by our method. Similarly to the case of still images, the compensated videos show reduced halos and higher constancy in flat regions. No temporal artifacts are observed when processing the frames independently.

\begin{figure}[t]
\centering
\includegraphics[width=3in]{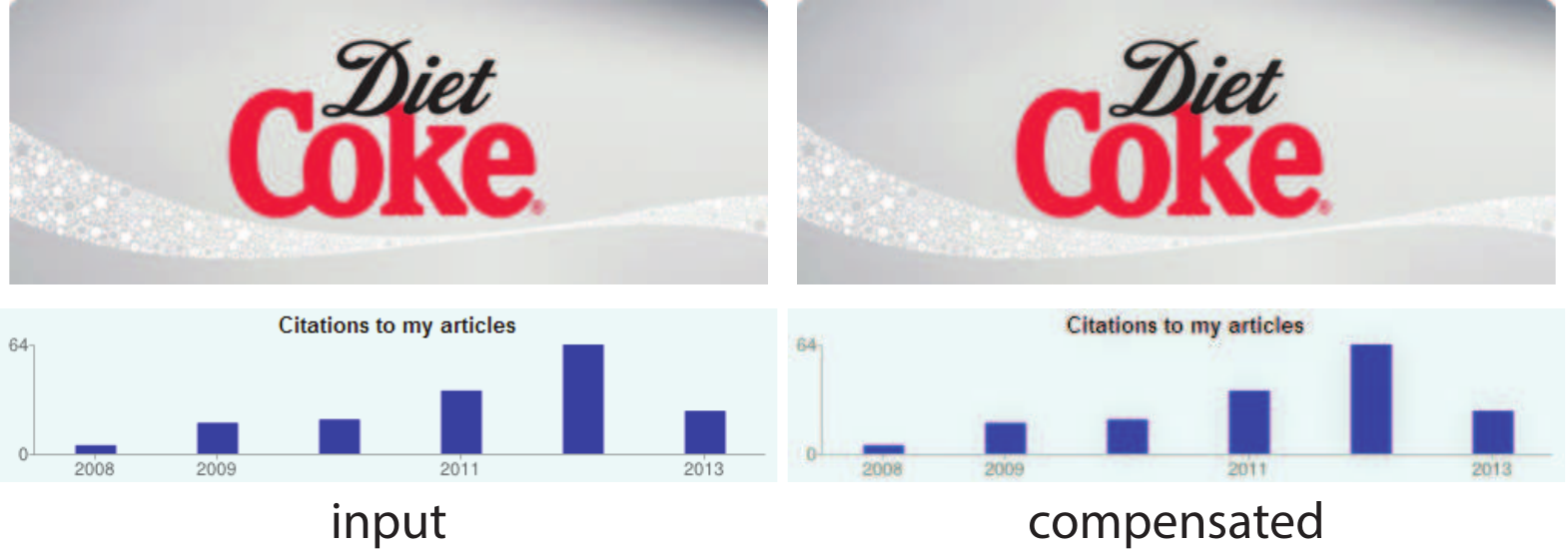}
\caption{Additional image domains benefiting our approach. The red Coke label shows halos which are reduced by our correction. Similarly, the false brightening around the Google Scholar bars is reduced. In all cases the laterally-compensated images give a softer appearance.}	
\label{fig:logo_tech}
\vspace{-0.27in}
\end{figure}

\vspace{-0.01in}
Logos and graphs are additional scenarios that can benefit from our compensation. As demonstrated in Figure~\ref{fig:logo_tech}, the halos around the red Coke text or the  Google Scholar bars are reduced in the laterally-compensated images which offer a softer appearance and less distracting features. Technical drawing, maps and computer games with small color-palette also exhibit Mach bands and false halos. We demonstrate the utility of our approach to such images in the supplemental material.



\vspace{-0.01in}
\bt{Web Pages.} A vast number of web pages, written in HTML or other script language, are composed of blocks of different functionality, such as advertisement banners and video players. This involves many high-contrast and sharp edges that produce halos and Mach bands effects. YouTube's screen-shot, found at the supplemental material, shows a bright halo appears around the video player's frame. In our compensated image this effect is reduced and the page appears calmer to the eye. Note that it is possible to apply our correction only to specific blocks, e.g., the player's frame but not process the video content.

\begin{figure}[t]
\centering
\includegraphics[width=1.9in]{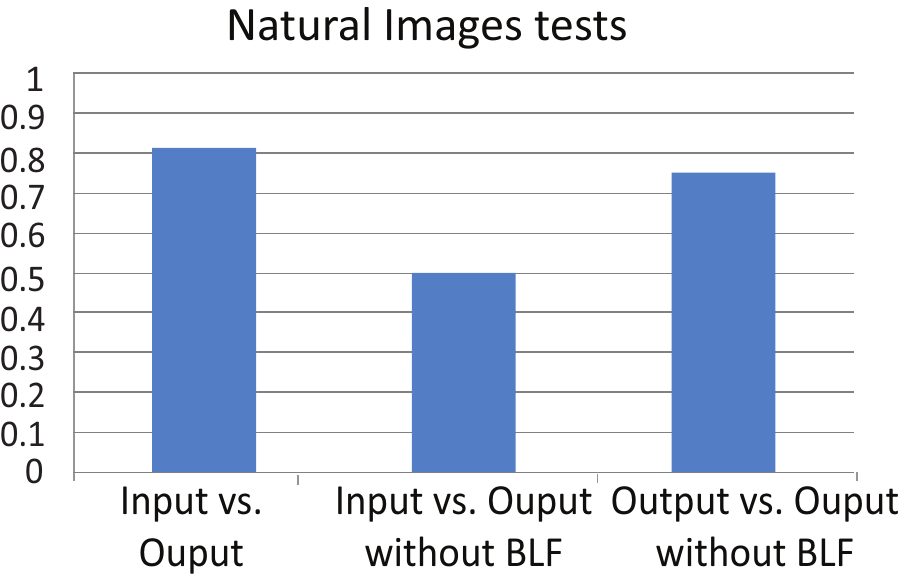}
\caption{Visual quality over natural images. Left bar shows the proportion of viewers who preferred the compensated image over the input. Centered bar shows the preference for compensation without the detail preservation step, and the right bar shows the preference between compensation with and without this step.}
\label{fig:details_exp}
\vspace{-0.2in}
\end{figure}
\begin{figure*}[t]
\centering
\includegraphics[width=6.3in]{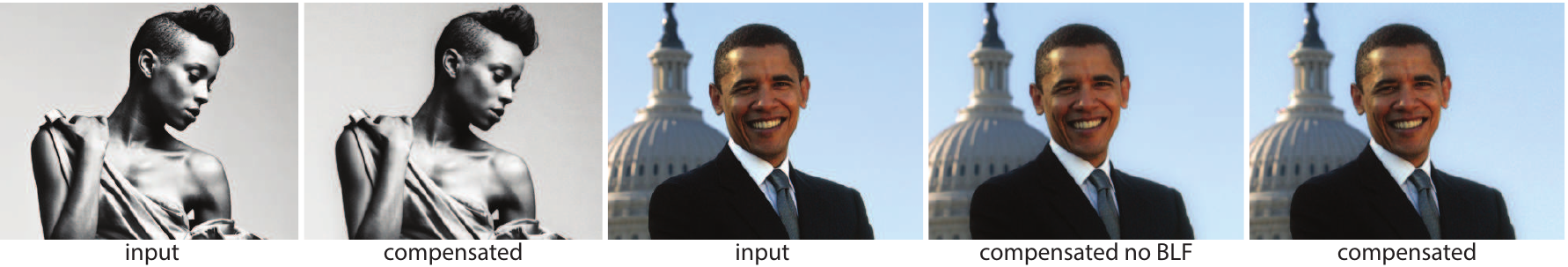}
\caption{Compensating natural images. False halos are seen around the strong edges, such as the woman's profile and Obama's outline. Second image from the right shows the compensated image without using the detail enhancement procedure.}
\label{fig:natural}
\vspace{-0.2in}
\end{figure*}

\vspace{-0.01in}
\bt{Natural Images.} As we acknowledged in Section~\ref{sec:introduction}, it is widely-accepted that the human visual system is adjusted to natural scenery. Nevertheless, we experimented in applying our method over this type of images. The images shown in Figure~\ref{fig:natural} contain high-contrast edges around the persons' outlines and around which halos are perceived. In these images the outline edges consist chiefly of a monotonic pixel variation (step-edges). Similarly to the other types of images we discussed above, the compensation reduces the halos and flattens the background. The softer appearance resulting from the idealized step edges may be preferable to some or serve as an artistic filter. In order to assess the visual impact of our method, we presented the viewers a natural images next to its compensation and asked them to specify ``which image looks better?".  The left bar in Figure~\ref{fig:details_exp} shows that the majority chose the compensated image.

\vspace{-0.01in}
The need for the detail preservation step, described in Section~\ref{sec:details}, is typically unnecessary in images that do not contain a considerable amount of fine details, e.g., cartoons, logos, maps, technical drawings etc. Natural images may contain such detail and hence we applied it only when processing natural images. Figure~\ref{fig:natural} demonstrates this step's ability to preserve fine-details such as the facial features. In order to quantify its utility we repeated the test above and compared the input to its compensation without this step as well as compared two compensated images, with and without the detail preservation step. The results, shown by the middle and right bars in Figure~\ref{fig:details_exp}, indicate that this step is needed for obtaining better visual quality.

\vspace{-0.15in}

\section{Conclusions}
\vspace{-0.05in}
We described a new algorithm that compensates for visual biases produced by the lateral inhibition mechanism in the retina. The method operates based on a perception-based model consisting of two well-studied physiological mechanisms: the compressed perception of light intensity and the neural inhibition process in the retina. In order to apply the method on color images, we derived an extension of the Hartline-Ratliff formula that accounts for the three types of cone cells in the human retina based on a chromatically-blind coupling.

\vspace{-0.01in}
We conducted a user study to evaluate the method's effectiveness as well as estimate its parameters. The tests confirm the method's ability to reduce the lateral biases in various types of images and its applicability to various tasks, namely, improving the comprehension of imaged values, e.g., medical data, aiding the perception of brightness constancy and reducing Mach bands and halos in different types of images such as logos, maps, cartoons, computer games, technical plans and natural photographs. The latter reinforces the theoretical attributions of these illusions to the lateral inhibition mechanism.

\vspace{-0.01in}

Our approach suffers from several drawbacks. As noted in Section~\ref{sec:method} it depends on the viewing distance and may not be applicable in situations where the viewing distance is unknown or varies considerably. A sonar or a different type of depth sensor can offer a solution in this respect. Similarly, the conversion from RGB to XYZ depends on the display and needs calibration. We should note that low-cost calibration devices are readily available nowadays. While the method reduces lateral biases in natural images, it is also introduces some blur that weaken the fine details. The detail-preserving step we described offers some remedy for this shortcoming. As we noted earlier, the lateral inhibition mechanism is a time-dependent process and our method assumes a steady viewing configuration. Hence, when moving the eye across the image our compensation is usually lost. Despite these limitations, we believe that our work makes practical contributions in many scenarios as well as hope that it will increase the awareness in the computer graphics and computational photography communities to the biases associated with the lateral inhibition mechanism.
\vspace{-0.01in}

As future work we intend to explore the modeling and compensation of other visual mechanisms in order to address additional types of visual illusions, such as ones associated with peripheral vision. While we demonstrated feasibility, as future work we intend to study more thoroughly the application of our approach in medical data.
\vspace{-0.1in}

\section* {Acknowledgements}

We thank Prof. Jacob Sosna and Dr. Chen Nadler for their useful feedback on our work. This work was funded by the Israeli Science Foundation (ISF) as well
as the European Research Council, ERC Starting Grant $337383$ "Fast-Filtering". 

\bibliographystyle{eg-alpha}
\renewcommand{\baselinestretch}{-0.6}
\bibliography{inbar-paper.bib}
\bt{Appendix:}


We extend here the Hartline-Ratliff model:
\begin{equation}
\label{eq:HR}
r_n = e_n - \sum_mk_{nm}r_m,
\vspace{-0.1in}
\end{equation}
(where $k_{nm} $ are the interaction coefficients of the inhibitory kernel) to account for lateral inhibition between the three types of cone cells in the human retina. The most general form of (linear) inhibition is given by
\begin{equation}
\label{eq:color_HR}
r_n^{c_n} = e_n^{c_n} - \sum_m k_{n m c_n c_m}r_m^{c_m},
\end{equation}
where $c_n, c_m\in \{ L,M,S\}$ denote the long-, medium-, and short-wavelength cell types, $n$ and $m$ index interacting neural cells and $e_n $ and $ r_n$ are the excitation and the inhibited excitation in the second level of the Hartline-Ratliff neural network respectively.

\begin{figure}[t]
\centering
\includegraphics[width=2in]{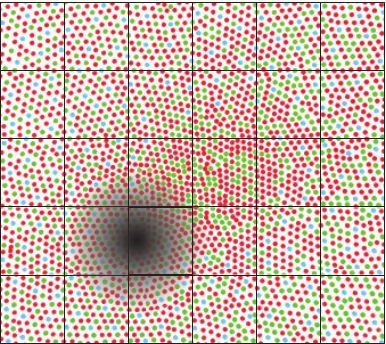}
\caption{Cone cells in the retina. Red, green and blue dots indicate the long-, medium- and short-wavelength cone types respectively. The values of a single inhibition kernel $\blur(\x-\x_n)$ around the $n$-th cone are shown in grey-scale. The grid-bins used in our derivation are outlined in black.}	
\label{fig:cones_in_retina}
\end{figure}

According to~\cite{boycott87,dacey96} the horizontal cells, which transmit inhibitory signal to bipolar cells, receive their input from cones cells regardless of  their type (H2 horizontal cells are indiscriminately connected to the long- and medium-wavelength cell types and H1 cells are connected indiscriminately to all three cell types). Hence, we set $ k_{n m c_n c_m}= k_{n m}$. Here again, we assume the lateral inhibition depends only on the relative distance between the cells and therefore $ k_{n m} = k(\x_n-\x_m) $, where $ \x_n $ and $ \x_m$ are the cells positions in the retina. Thus, we rewrite Eq.~\eqref{eq:color_HR} as
\begin{equation}
\label{eq:model_cones}
r_n^{c_n} = e_n^{c_n} - \sum_mk(\x_n-\x_m) r_m^{c_m}.
\end{equation}
We divide the retinal plane into a grid of square bins $ G_{\x}$, as illustrated in Figure~\ref{fig:cones_in_retina}, where the bins are indexed by their center point $\x$. We denote the set of cells inside the $\x$-th bin and of type $c \in \{L,M,S\} $ by $ G^c_\x = \{i : \x_i \in G_{\x}, c_n = c\}$. Assuming the cone cells are uniformly distributed in the retina, the occurrence of a particular cell type should be equal in each grid-bin, i.e., $| G^c_\x |=| G^c |$ for every cell type, $c \in \{L,M,S\} $.

We define the average response and input stimulus in each bin respectively by
\begin{equation}
\perc^c(\x) = \frac{1}{|G^c|} \sum_{n \in G^c_\x} r_n^{c_n},  \:\: \mathrm{and}\:\: \ems^c(\x) = \frac{1}{|G^c|} \sum_{n \in G^c_\x} e_n^{c_n}.
\end{equation}
Next, we assume that the grid is fine enough and use the bin center coordinates $\x$ to approximate the location of the cells inside bin, i.e., $\x_n \approx \x$ for all the cells inside the bin $G_x$. Based on this approximation and the kernel smoothness we get that $ k(\x_n-\x_m) \approx k(\x-\x_m)$. We use the latter to express Eq.~\eqref{eq:model_cones} in grid representation using the cell-average response and stimulus functions by
\begin{equation}
\label{eq:approx}
\begin{aligned}
\perc^c(\x) &= \ems^c(\x) - \frac{1}{|G^c|} \sum_{n \in G_\x^c} \sum_mk(\x_n-\x_m) r_m^{c_m} \\
& \approx \ems^c(\x) - \sum_mk(\x-\x_m) r_m^{c_m}.
\end{aligned}
\end{equation}
This equation is obtained by summing both sides of Eq.~\eqref{eq:model_cones} at the $\x$-th bin and divide by $|G^c|$. We also used $ |G^c|^{-1} \sum_{n \in G_\x^c}  = 1 $. The last term in Eq.~\eqref{eq:approx} is written more explicitly as
\begin{equation}
\sum_m k(\x-\x_m) r_m^{c_m} = \!\!\!\! \sum_{c' \in \{L,M,S\}} \!\! \sum_{\y} \sum_{\x_m \in G^{c'}_\y} k(\x-\x_m) r_m^{c_m},
\end{equation}
We repeat the same approximation as above and assume that $\x_m \approx \y$ for the cells inside the bin centered around $\y$ and use $  k(\x-\x_m) \approx k(\x-\y)$ to obtain
\begin{equation}
\begin{aligned}
& \sum_{c' \in \{L,M,S\}}\sum_{\y} \sum_{\x_m \in G^{c'}_\y} k(\x-\y) r_m^{c_m} \\
= &\!\!\!\!\!\!\sum_{c' \in \{L,M,S\}}\!\!\sum_{\y} k(\x-\y)   \frac{ |G^{c'}|}{|G^{c'}|}  \sum_{\x_m \in G^{c'}_\y} r_m^{c_m} \\
= & \sum_{\y} k(\x-\y)  \!\!\!\!\!\!\sum_{c' \in \{L,M,S\}} \!\!\!\!\!\!|G^{c'}| \perc^{c'}(\y).
\end{aligned}
\end{equation}

Now, by taking the grid bins to be infinitesimally small, the number of grid centers $\y$ grows and the number of cone cells within each bin decreases. Thus, we obtain the following Riemann integral
\begin{equation}
\begin{aligned}
 & \sum_{\y} k(\x-\y) \!\!\!\!\!\!\!\! \sum_{c' \in \{L,M,S\} } \!\!\!\!\!\!\!\!|G^{c'}| \perc^{c'}(\y) \rightarrow  \int_{\y}  k(\x-\y) \big( \!\!\!\!\!\!\!\!\sum_{c' \in \{L,M,S\} }  \!\!\!\!\!\!\!\! p^{c'} \perc^{c'}(\y) \big) d\y  \\
 & =  \int_{\y} k(\x-\y)  \perc(\y) d\y = (k * \perc) (\x),
 \end{aligned}
\end{equation}
where $\perc(\y) $ denotes the total perceived excitation in the retina, produced by all three type of cone cells. The three scalars $p^L, p^M $and $p^S$ correspond to the proportion of each cell type in the retina. Since there are typically 50\% more long-wavelength cones than medium-wavelength cells~\cite{carroll02} and the short-wavelength cones constitute about ten percent of the total cones population, we define the total excitation by
\begin{equation}
\perc(\x) = \sfrac{3}{2}\perc'_L(\x) + \perc'_M(\x) + \perc'_S(\x)/4.
\end{equation}
Finally, we obtain that
\begin{equation}
\perc^c(\x) = \ems^c(\x) -  (k * \perc) (\x)
\end{equation}
which we use in Eq. 6 in the paper.

\end{document}